\documentclass[runningheads]{llncs}

\usepackage{makeidx}
\usepackage{graphicx}
\usepackage{amsmath,amssymb} 
\usepackage{color}

\usepackage{siunitx}
\usepackage{subcaption}
\usepackage[labelfont=bf]{caption}
\usepackage{ctable}
\setupctable{captionskip=5pt}
\usepackage{rotating}
\usepackage{multirow}
\usepackage{breakcites}
\newcommand{\rotatedlabel}[1]{\begin{sideways}#1\end{sideways}}

\usepackage{xcolor}

\usepackage{comment}
\newcommand{\multilinecell}[2][c]{%
  \begin{tabular}[#1]{@{}c@{}}#2\end{tabular}}
\usepackage{bibentry}

\usepackage[pagebackref=true,breaklinks=true,colorlinks,bookmarks=false]{hyperref}

\usepackage[capitalize]{cleveref}

\crefname{section}{Sec.}{Sec.}

\newcommand{\bst}[1]{$\mathbf{#1}$} 
\newcommand{\myparagraph}[1]{\vspace{0.5em}\noindent\textbf{#1}}
\newcommand{\mapr}{\text{AP}}
\newcommand{\iou}{\text{IoU}}
\newcommand{\miou}{\text{IoU}}
\newcommand{\iiou}{\text{iIoU}}
\newcommand{\miiou}{\text{iIoU}}
\newcommand{\coarselevel}{\text{category}}
\newcommand{\finelevel}{\text{class}}

\usepackage{xspace}
\makeatletter
\@ifundefined{onedot}{
    \DeclareRobustCommand\onedot{\futurelet\@let@token\@onedot} %
    \def\@onedot{\ifx\@let@token.\else.\null\fi\xspace} %
    \def\eg{\emph{e.g}\onedot}  %
    \def\ie{\emph{i.e}\onedot}  %
    \def\cf{\emph{c.f}\onedot}  %
}
\makeatother

\usepackage{lineno}

\begin{document}
    \pagestyle{headings}
    \mainmatter

    \def\GCPR16SubNumber{13}

    \title{Pixel-level Encoding and Depth Layering for Instance-level Semantic Labeling\vspace{-0.5em}}

    \titlerunning{ Pixel Encoding and Depth Layering for Instance-level Semantic Labeling }
    \authorrunning{ Jonas Uhrig, Marius Cordts, Uwe Franke, Thomas Brox }
    \author{Jonas Uhrig$^{1,2}$, Marius Cordts$^{1,3}$, Uwe Franke$^{1}$, Thomas Brox$^{2}$\vspace{-0.5em}}
    \institute{$^1$Daimler AG R\&D, $^2$University of Freiburg, $^3$TU Darmstadt\\\texttt{jonas.uhrig@daimler.com}\vspace{-0.5em}}

    \maketitle

    \begin{abstract}
        Recent approaches for instance-aware semantic labeling have augmented
        convolutional neural networks (CNNs) with complex multi-task architectures or
        computationally expensive graphical
        models. We present a method that leverages a fully convolutional
        network (FCN) to predict semantic labels,
        depth and an instance-based encoding using each pixel's direction towards
        its corresponding instance center. Subsequently,
        we apply low-level computer vision techniques to generate state-of-the-art
        instance segmentation on the street scene datasets KITTI
        and Cityscapes. Our approach outperforms existing works by a
        large margin and can additionally predict absolute distances of
        individual instances from a monocular image as well as a
         pixel-level semantic labeling.\vspace{-0.5em}
    \end{abstract}
    \vspace{-0.5em}

    \begin{figure}[b]
    \centering
    \begin{subfigure}{0.332\textwidth}
        \includegraphics[width=\textwidth]{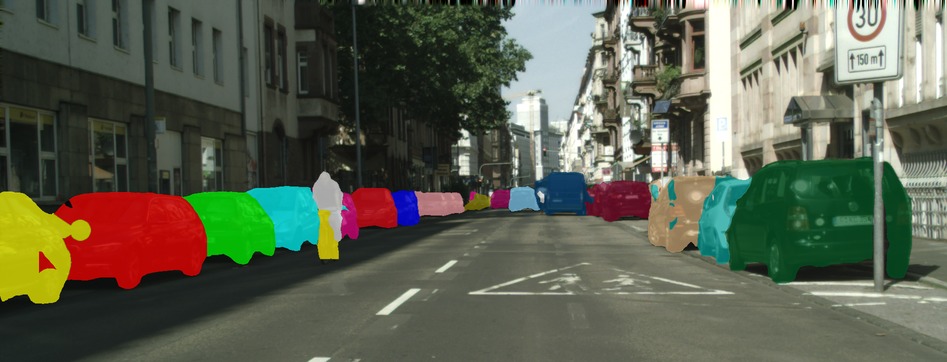}
    \end{subfigure}\hspace{0.5pt}%
    \begin{subfigure}{0.332\textwidth}
        \includegraphics[width=\textwidth]{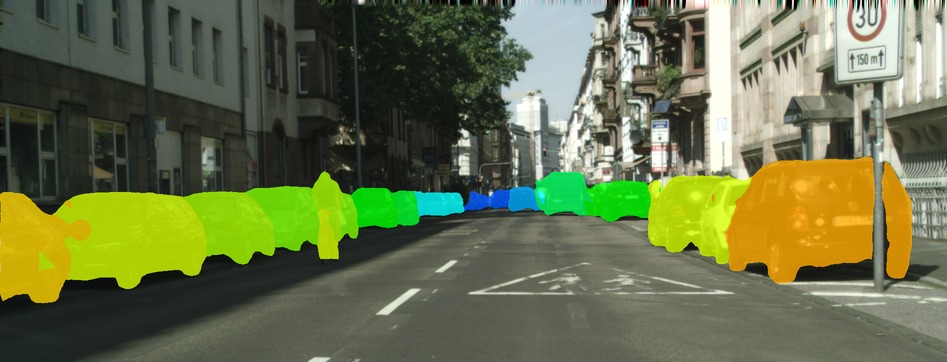}
    \end{subfigure}\hspace{0.5pt}%
    \begin{subfigure}{0.332\textwidth}
        \includegraphics[width=\textwidth]{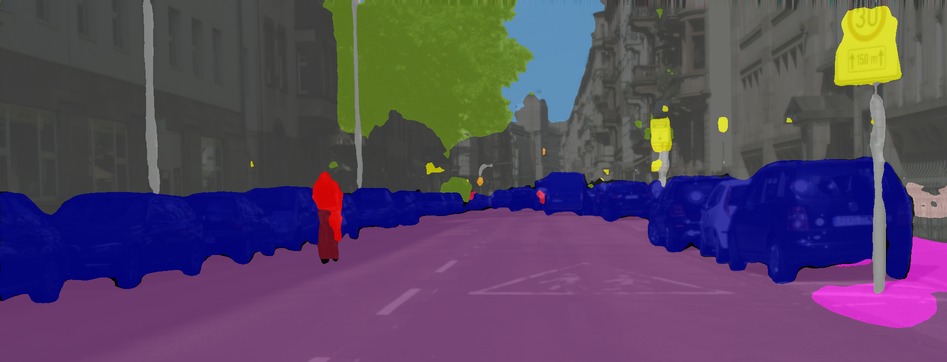}
    \end{subfigure}%
    \caption{Example scene representation as obtained by our method: instance segmentation,
                monocular depth estimation, and pixel-level semantic labeling.\vspace{-0.5em}}

    \label{fig:titlefigure}
    \end{figure}

    \section{Introduction}
    \label{sec:introduction}

    The task of visual semantic scene understanding is mainly tackled from two
    opposing facets: pixel-level semantic labeling
    \cite{ChenPapandreou2014,FCN2015,Papandreou2015} and bounding-box
    object detection \cite{girshick15fastrcnn,girshick2014rcnn,YOLO2015,ren2015faster}.
    The first assigns each pixel in an image with a semantic label segmenting
    the semantically connected regions in the scene.
    Such approaches work well with non-compact (\emph{background})
    classes such as buildings or ground, yet they do not
    distinguish individual object instances.
    Object detection aims to find all individual instances in the scene and
    describes them via bounding boxes. Therefore, the latter provides a rather
    coarse localization and is restricted to compact (\emph{object}) classes
    such as cars or humans.

    Recently, instance-level semantic labeling gained increasing interest
    \cite{Dai2015,Liang2015,Urtasun2015a,Urtasun2015b}. This task is at the
    intersection of both challenges. The aim is to
    combine the detection task with instance segmentation. Such a
    representation allows for a precise localization, which in turn enables
    better scene understanding. Especially in the domain of robotics and
    autonomous vehicles, instance-level semantic segmentation enables an
    explicit occlusion reasoning, precise object tracking and motion
    estimation, as well as behavior modeling and prediction.

    Most state-of-the-art methods build upon a fully convolutional
    network (FCN) \cite{FCN2015}. Recent approaches typically add
    post-processing, for example, based on conditional
    random fields (CRFs)
    \cite{Urtasun2015a,Urtasun2015b}. Other methods score
    region proposals for instance segmentation
    \cite{Dai2015FeatureMasking,Hariharan2014} or object detection
    \cite{girshick15fastrcnn,girshick2014rcnn,YOLO2015,ren2015faster}, or use a
    multi-stage neural network for these tasks \cite{Dai2015,Liang2015}.

\begin{figure}[t]
    \centering
    \includegraphics[width=\textwidth]{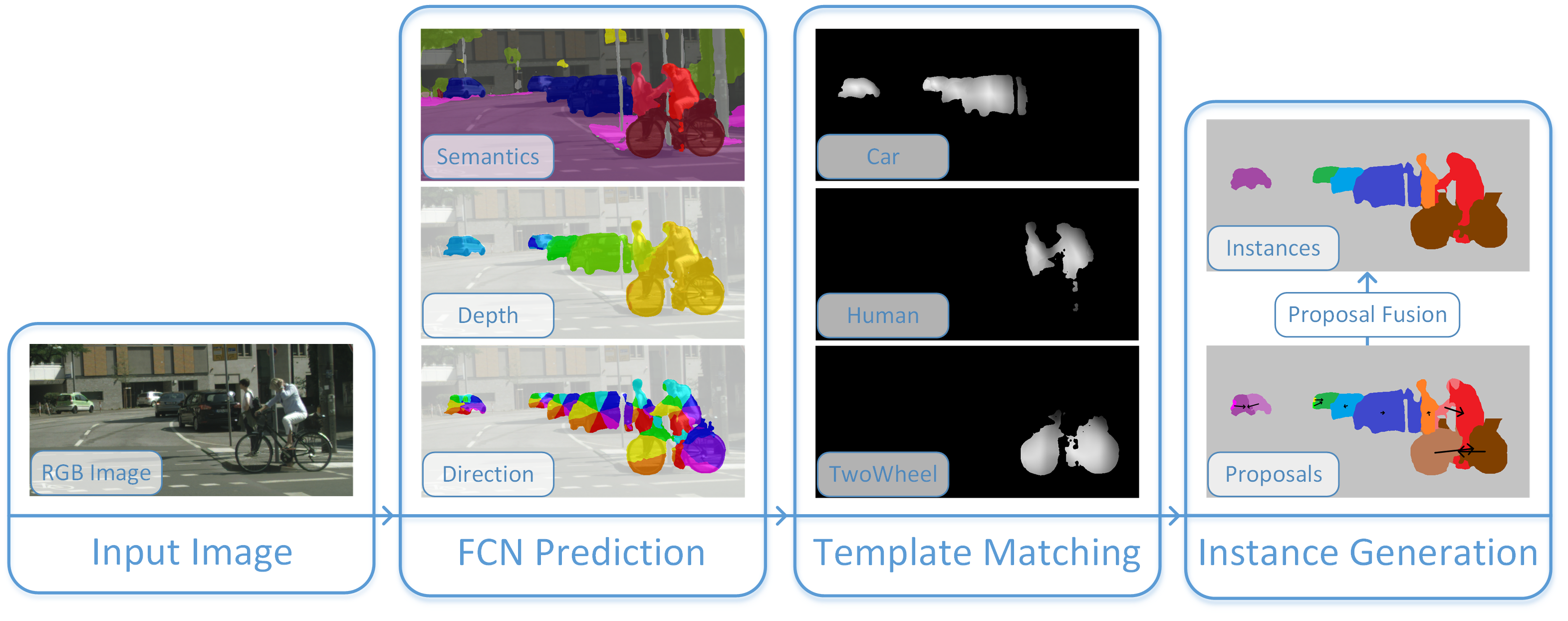}
    \caption{From a single image, we predict \num{3} FCN outputs:
                 semantics, depth, and instance center direction.
                 Those are used to compute template matching
                 score maps for semantic categories.
                 Using these, we locate and generate instance proposals
                 and fuse them to obtain our instance segmentation.\vspace{-0.5em}}
    \label{fig:methodsummary}
    \end{figure}

    In this work, we focus on street scene understanding and use a single
    monocular image to simultaneously obtain a holistic scene representation,
    consisting of a pixel-level semantic labeling, an instance-level
    segmentation of traffic participants, and a 3D depth estimation for
    each instance. We leverage an FCN that yields powerful
    pixel-level cues consisting of three output channels: a semantic class,
    the direction to the object center (where applicable) and the object
    distance (where applicable).
    Scene understanding is mainly due to the network and post-processing with standard
    computer vision methods is sufficient to obtain a detailed representation
    of an instance-aware semantic segmentation, \cf \cref{fig:titlefigure,fig:methodsummary}.
    Our method significantly outperforms state-of-the-art methods on the
    street scene datasets KITTI \cite{KITTI2012} and
    Cityscapes \cite{Cordts2015}.

    \section{Related Work}
    \label{sec:relatedwork}

    For the task of instance-level semantic labeling, there exist two major
    lines of research. The first leverages an over-complete set of
    object proposals that are either rejected, classified as an instance of a
    certain semantic class, and refined to
    obtain an instance segmentation. Common to all such methods is that the
    performance is depending on the quality of these proposals, since they
    cannot recover from missing instances in the proposal stage. Generally,
    such approaches tend to be slow since all
    proposals must be classified individually.
    These properties cause inaccurate proposals to limit the performance of such
    methods~\cite{Cordts2015,Hosang2015Pami}. Our method belongs to the
    category of proposal-free methods, where the
    segmentation and the semantic class of object instances are inferred jointly.

    \myparagraph{Proposal-based instance segmentation.}
    Driven by the success of deep learning based object detectors
    such as R-CNN~\cite{girshick2014rcnn} or its
    variants~\cite{girshick15fastrcnn,ren2015faster,He2015}, recent methods
    rely on these detections for instance segmentation. Either the underlying
    region proposals, such as MCG~\cite{Arbelaez2014}, are directly used as
    instance segments~\cite{Cordts2015,Dai2015FeatureMasking,Hariharan2014},
    or the bounding boxes are refined to obtain instance
    masks~\cite{Chen2015,Hariharan2015}. Instead of bounding boxes, \cite{objcut2005}
    uses a layered pictorial structure (LPS) model, where shape exemplars for
    object parts are mapped to the image in a probabilistic way. This yields an
    initial proposal for the object's pose and shape, which is refined using
    appearance cues. Using a bank of object detectors as
    proposals, \cite{yang2012layered} infers the instance masks via occlusion
    reasoning based on discrete depth layers. In \cite{Tighe2014}, pixel-level
    semantic labels are used to score object candidates and vice versa in an
    alternating fashion, while also reasoning about occlusions and scene
    geometry. Based on proposals that form a segmentation tree, an energy
    function is constructed
    in~\cite{Silberman2014} and its solution yields the instance segmentation.

    Recently, \cite{Dai2015} extended the R-CNN for instance
    segmentation with a multi-task network cascade. A fully convolutional
    network combined with three classification
    stages produces bounding-box proposals, refines these to
    segments, and ranks them to obtain the final instance-level
    labeling. They achieve excellent performance on PASCAL
    VOC~\cite{Pascal2010} and MS COCO~\cite{MSCoco2014}.

    \myparagraph{Proposal-free instance segmentation.}
    Pixel-level semantic labeling based on neural networks
    has been very
    successful~\cite{ChenPapandreou2014,Rother2015,FCN2015,Yu2016,ZhengJayasumana2015}.
    This triggered interest in casting also instance segmentation
    directly as a pixel labeling task.
    In \cite{Ronneberger2015}, the network predicts for each
    pixel, whether it lies on an object boundary or not, however, requiring a rather delicate training.
    Using a long short-term memory (LSTM) network~\cite{LSTM1997},
    instance segmentations can be sequentially sampled~\cite{Romera2015}.

    In \cite{Urtasun2015a,Urtasun2015b}, instances are encoded
    via numbers that are further constrained to encode relative depth ordering in order
    to prevent arbitrary assignments. An FCN predicts these IDs at
    each pixel and a subsequent Markov Random Field (MRF) improves these
    predictions and enforces consistency. However, such a method is limited to
    scenes, where a clear depth ordering is present, \eg a single row of parking cars,
    and the maximum number of instances is rather low.

    The proposal-free network (PFN)~\cite{Liang2015} is a CNN that yields a
    pixel-level semantic labeling, the number of instances in the scene, and
    for each pixel the parameters of a corresponding instance bounding box.
    Based on these predictions, instances are obtained by clustering.
    The network has a fairly complex architecture with many interleaved
    building blocks, making training quite tricky. Further, the overall
    performance highly depends on the correct prediction of the number of
    instances in the scene. In street scenes, there can be
    hundreds of instances per image~\cite{Cordts2015}. Thus, the number of
    training samples per number of instances is low, mistakes in their
    estimation can be critical, and the available cues for clustering
    might not correlate with the estimated number of instances.

    In this work, we focus on urban street scenes.
    Besides each pixel's semantic class, our network estimates
    an absolute depth, which is particularly useful for instance separation
    in street scenes. We encode instances on a pixel-level by the direction
    towards their center point. This representation is independent of the
    number of instances per image and provides strong
    signals at the instance boundaries.

    \section{Method}
    \label{sec:method}

    \subsection{FCN Feature Representation}
    \label{subsec:fcnoutputrepresentation}
    Our network extends the FCN-8s model~\cite{FCN2015} with
    three output channels that together facilitate instance
    segmentation. All channels are jointly trained as pixel-wise discrete
    labeling tasks using standard cross-entropy losses. Our proposed representation
    consists of (1) a semantic channel that drives the instance classification,
    (2) a depth channel to incorporate scale and support instance separation, and
    (3) a 2D geometric channel to facilitate instance detection and segmentation.

    We chose the upscaling part of our FCN such that we can easily change
    the number of classes for each of the three proposed channels without
    re-initializing all upsampling layers. To this end, after the largest
    downsampling factor is reached, we use Deconvolution layers together
    with skip layers~\cite{FCN2015} to produce a representation of $\frac{1}{8}$
    of the input resolution with a depth of \num{100} throughout all intermediate
    layers. The number of channels of this abstract representation is then
    reduced through $1\!\times\!1$ convolutions to the
    proposed semantic, depth, and instance
    center channels. To reach full input resolution, bilinear
    upsampling is applied, followed by a separate cross-entropy loss for
    each of our three output channels.

    \begin{figure}[t]
    \begin{subfigure}{0.32\textwidth}
        \includegraphics[width=\textwidth]{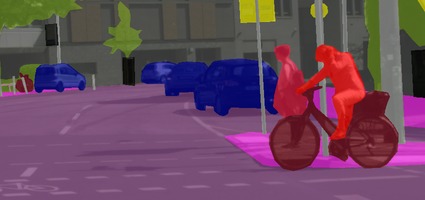}
        \caption{Semantic label.}
        \label{fig:semGT}
    \end{subfigure}
    \begin{subfigure}{0.32\textwidth}
        \includegraphics[width=\textwidth]{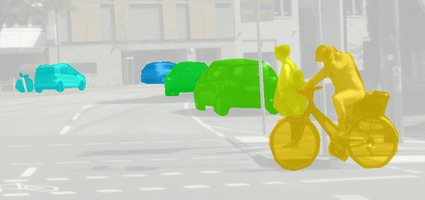}
        \caption{Depth class.}
        \label{fig:depthGT}
    \end{subfigure}
    \begin{subfigure}{0.32\textwidth}
        \includegraphics[width=\textwidth]{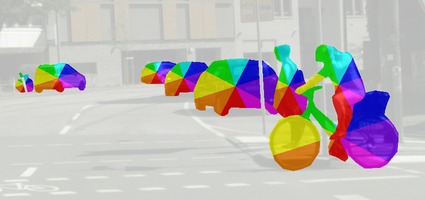}
        \caption{Instance direction.}
        \label{fig:dirGT}
    \end{subfigure}
    \caption{Ground truth examples of our three proposed FCN channels. Color
             overlay (a) as suggested by \cite{Cordts2015}, (b) represents
             depth per object from red (close) to blue (distant), (c)
             represents directions towards corresponding instance centers.\vspace{-0.5em}}

    \label{fig:semanticdepthexamples}
    \end{figure}

    \myparagraph{Semantics.}
    To cope with different semantic classes, we
    predict a semantic label for each input pixel, \cf \cref{fig:semGT}.
    These predictions are particularly important as they are the only
    source of semantic information in our approach. Further, the predicted
    semantic labels allow us to separate objects from background as well as
    objects of different classes from each other.

    \myparagraph{Depth.}
    Urban street scenes typically contain
    objects at various distances~\cite{Cordts2015}. To guide the post-processing
    in terms of objects at different scales, we predict a depth label
    for each object pixel.
    We assign all pixels within an object instance to a constant depth value,
    e.g. the median over noisy measurements or the center of a 3D bounding box,
    \cf \cref{fig:depthGT}. These depth estimates also support instance separation, which
    becomes apparent when considering a row of parking cars, where the depth delta between
    neighboring cars is a full car length instead of a few centimeters in continuous space.
    The depth values are discretized into a set of classes
    so that close objects have a finer depth resolution than distant objects.

    \myparagraph{Direction.}
    Object instances are defined by their boundary and class. Therefore, it
    seems natural to train an FCN model to directly predict boundary pixels.
    However, those boundaries represent a very
    delicate signal~\cite{amfm_pami2011} as they have a width of only one pixel,
    and a single erroneously labeled pixel in the training data has a much
    higher impact compared to a region-based representation.

    We introduce a class-based representation which implicitly combines
    information about an instance's boundary with the location of its
    visible center. For each object pixel we compute the direction
    towards its corresponding center and discretize this angle to a
    set of classes, \cf \cref{fig:dirGT}. This information is easier
    to grasp within a local region and is tailored for an FCN's
    capability to predict pixel-wise labels.
    Especially for pixels on the boundary between neighboring objects, our
    representation clearly separates the instances
    as predictions have nearly opposite directions. Since we predict the
    center of the visible area of an object and not its physical
    center, we can handle most types of occlusions very well.
    Furthermore, instance centers
    have a distinct pattern, \cf \cref{fig:dirGT},
    which we exploit by applying template matching, as described in
    \cref{subsec:templatematching}.
    Even though our proposed representation does not directly yield instance IDs,
    it is well defined even for an arbitrary number
    of instances per image.

    To obtain an accurate direction estimation for each pixel,
    we assign the average direction by weighting all direction vectors with
    their respective FCN score (after softmax normalization). This allows us
    to recover a continuous direction estimation from the few discretized
    classes.

    \subsection{Template Matching}
    \label{subsec:templatematching}
    To extract instance centers, we propose template matching on the direction
    predictions, where templates are rectangular and contain the
    distinct pattern visible in \cref{fig:dirGT}. We adjust the template's
    aspect ratio depending on its semantic class, so we can better
    distinguish between pedestrians and vehicles. In order to detect also
    distant objects with consistent matching scores, we scale the size
    of the templates depending on the predicted depth class.

    To reduce induced errors from confusions between objects of similar semantic classes,
    we combine multiple semantic classes into the categories \textit{human}, \textit{car},
    \textit{large vehicle}, and \textit{two wheeler}.

    Normalized cross-correlation (NCC) is used to produce a score map for each
    category by correlating all pixels with their respective template. These
    maps indicate the likelihood of pixels being an instance center, \cf \cref{fig:methodsummary}.
    In the following, we predict instances for each category separately. After all instances
    are found, we assign them the majority semantic class label.

    \subsection{Instance Generation}
    \label{subsec:instancegeneration}

    \myparagraph{Instance Centers.}
    To determine instance locations, we iteratively find maxima
    in the generated template matching score maps via non-maximum
    suppression within an area that equals the template size. This helps avoid
    multiple detections of the same instance while incorporating
    typical object sizes. Those maxima represent our
    \textit{temporary instance centers}, which are
    refined and merged in the following steps.

    \myparagraph{Instance Proposals.}
    Each pixel with a predicted direction from the FCN is assigned to the closest
    temporary instance center where the relative location and predicted
    direction agree.
    Joining all assigned pixels per instance hypothesis
    yields a set of \textit{instance proposals}.

    \myparagraph{Proposal Fusion.}
    \label{subsubsec:instancefusion}
    Elongated objects and erroneous depth predictions cause an over-segmentation
    of the instances.
    Thus, we refine the generated instances by accumulating estimated directions
    within each proposal. When interpreting direction predictions as vectors,
    they typically compensate each other within
    instance proposals that represent a complete instance, \ie there are as many
    predictions pointing both left and right.
    However, incomplete instance proposals are biased to a certain
    direction. If there is a neighboring instance candidate with matching
    semantic class and depth in the direction of this bias, the two proposals
    are fused.

    To the remaining instances we assign the
    average depth and the most frequent semantic class label within the
    region.
    Further, we merge our instance prediction with the pixel-level semantic
    labeling channel of the FCN by assigning the argmax semantic label to all
    non-instance pixels. Overall, we obtain a consistent scene representation,
    consisting of object instances paired with depth
    estimates and pixel-level labels for background classes.

    \section{Experiments}
    \label{sec:experiments}

    \subsection{Datasets and Metrics}
    \label{subsec:datasetsandmetrics}
    We evaluated our approach on the KITTI object detection dataset\cite{KITTI2012} extended with
    instance-level segmentations \cite{Chen2014,Urtasun2015b} as well as
    Cityscapes \cite{Cordts2015}. Both datasets provide pixel-level annotations
    for semantic classes and instances, as well as depth information, which is
    essential for our approach. For the ground truth
    instance depths we used the centers of their 3D bounding box annotation in KITTI and
    the median disparity for each instance in Cityscapes based on
    the provided disparity maps.
    We used the official splits for training, validation and test sets.

    We evaluated the segmentation based on the metrics proposed in \cite{Urtasun2015b} and
    \cite{Cordts2015}. To evaluate the depth prediction,
    we computed the mean absolute error (MAE), the
    root mean squared error (RMSE), the absolute relative difference (ARD),
    and the relative inlier ratios ($\delta_1$, $\delta_2$, $\delta_3$) for thresholds
    $\delta_i = 1.25^i$ \cite{Wang2015}. These metrics are computed on an instance level using the
    depths in meters. We only considered instances that overlap by more than
    \SI{50}{\percent} with the ground truth.

    \subsection{Network Details}
    For Cityscapes, we used the \num{19} semantic classes and combined the \num{8}
    object classes into \num{4} categories (\textit{car}, \textit{human}, \textit{two-wheeler},
    and \textit{large vehicle}). For KITTI, only \textit{car} instance
    segmentations are available.
    For both datasets, we used \num{19} depth classes and an explicit
    class for background. We chose ranges for each depth class and template
    sizes differently for each dataset
    to account for different characteristics of present objects
    and used camera settings~\cite{Cordts2015}.
    This is necessary as distances and semantic classes of objects
    differ remarkably. Details are provided in the supplementary material.
    The instance directions were split into \num{8} equal
    parts, each covering an angle of \SI{45}{\degree} for both datasets.

    We use the 8-stride version of an FCN, which is initialized using the
    ImageNet dataset~\cite{ImageNet}. After initializing the upsampling layers
    randomly, we fine-tune the network on KITTI and Cityscapes to obtain
    all three output channels.

    \setlength{\tabcolsep}{3pt}
    \ctable[
        caption = {Evaluation of our variants on KITTI \textit{val} (top)
                   and comparison with baselines (\textit{Best}~\cite{Urtasun2015a}/\cite{Urtasun2015b}) on KITTI \textit{test} (bottom) using metrics from
                   \cite{Urtasun2015b}. For AvgFP and AvgFN lower is better,
                   all other numbers are in percent and larger is better. \textit{Mix}~\cite{Urtasun2015b} shows the best results per metric from all baseline variants.},
        label   = {tab:kittiVariantsZhang},
        pos     = {tb},
        doinside= \scriptsize,
        width   = \textwidth
    ]{lXcccccccccc}{}{
            \FL
            Method                   & Set  & IoU        & MWCov      & MUCov      & AvgPr      & AvgRe      & AvgFP       & AvgFN       & InsPr      & InsRe      & InsF1      \ML
            Ours-D-F                 & val  & $79.4$     & $41.5$     & $43.4$     & \bst{92.8} & $54.4$     & $0.042$     & $1.33$      & $16.6$     & $29.8$     & $21.4$     \NN
            Ours-F                   & val  & \bst{82.2} & $35.9$     & $35.7$     & $83.6$     & \bst{86.7} & $0.158$     & \bst{0.100} & $31.4$     & $69.5$     & $43.3$     \NN
            Ours-D                   & val  & $79.6$     & \bst{82.4} & \bst{79.9} & $89.9$     & $54.6$     & \bst{0.017} & $1.33$      & \bst{96.0} & $42.8$     & $59.2$     \NN
            Ours                     & val  & \bst{82.2} & $80.7$     & $76.3$     & $83.7$     & \bst{86.7} & $0.100$     & \bst{0.100} & $91.8$     & \bst{82.3} & \bst{86.8} \ML
            Best \cite{Urtasun2015a} & test & $77.4$     & $67.0$     & $49.8$     & $82.0$     & $61.3$     & $0.479$     & $0.840$     & $48.9$     & $43.8$     & $46.2$     \NN
            Best \cite{Urtasun2015b} & test & $77.0$     & $69.7$     & $51.8$     & $83.9$     & $57.5$     & $0.375$     & $1.139$     & $65.3$     & $50.0$     & $56.6$     \NN
            Mix  \cite{Urtasun2015b} & test & $77.6$     & $69.7$     & $53.9$     & $83.9$     & $63.4$     & $0.354$     & $0.618$     & $65.3$     & $52.2$     & $56.6$     \NN
            Ours                     & test & \bst{84.1} & \bst{79.7} & \bst{75.8} & \bst{85.6} & \bst{82.0} & \bst{0.201} & \bst{0.159} & \bst{86.3} & \bst{74.1} & \bst{79.7} \LL
    }

    \begin{figure}[t]
    \centering
    \begin{subfigure}{0.33\textwidth}
        \includegraphics[width=\textwidth]{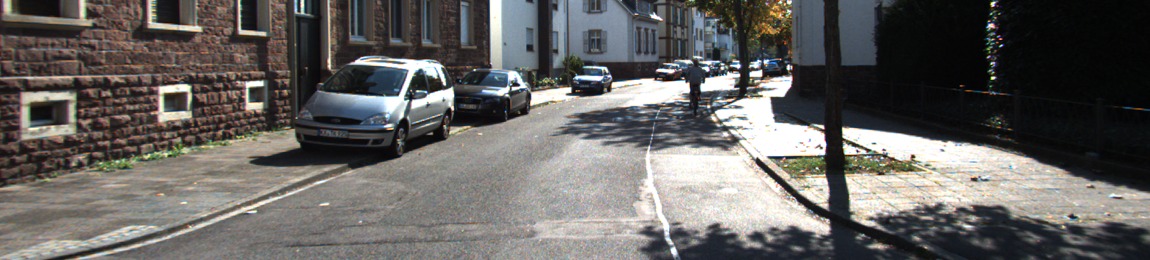}
    \end{subfigure}\hspace{1pt}%
    \begin{subfigure}{0.33\textwidth}
        \includegraphics[width=\textwidth]{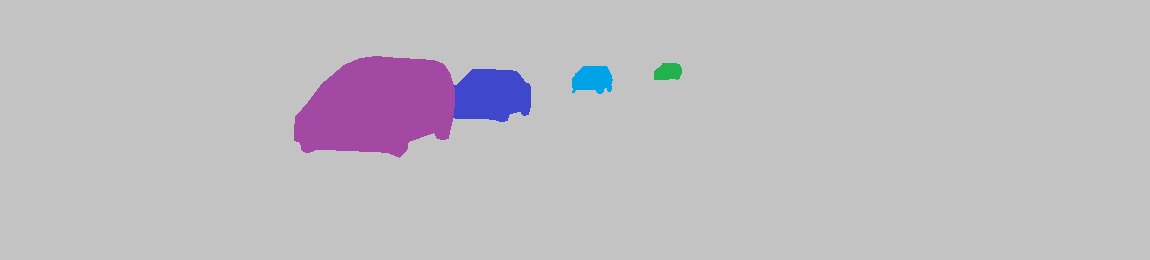}
    \end{subfigure}\hspace{1pt}%
    \begin{subfigure}{0.33\textwidth}
        \includegraphics[width=\textwidth]{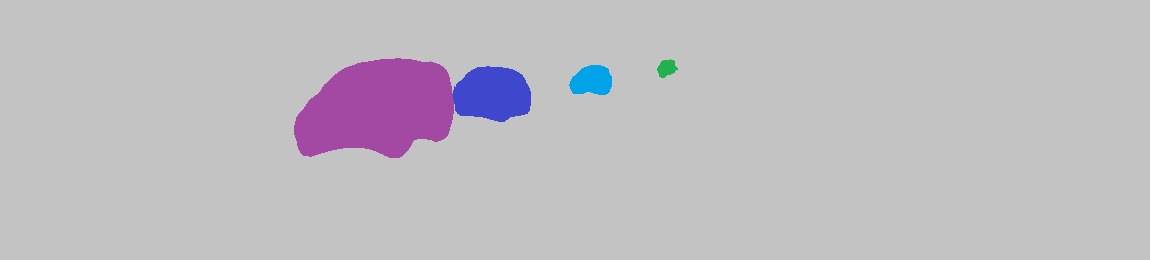}
    \end{subfigure}\\
    \begin{subfigure}{0.33\textwidth}
        \includegraphics[width=\textwidth]{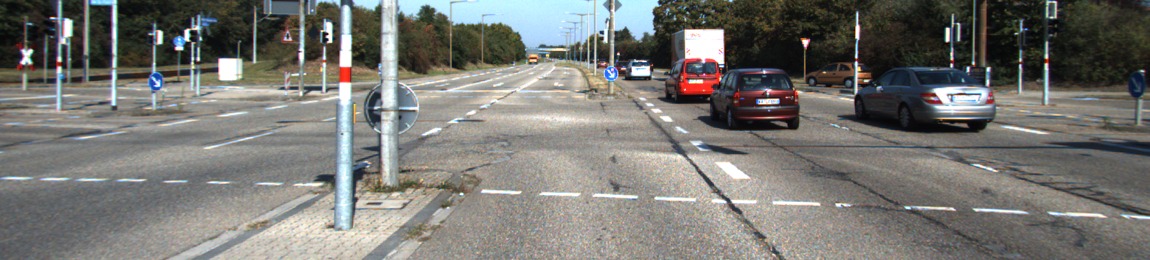}
    \end{subfigure}\hspace{1pt}%
    \begin{subfigure}{0.33\textwidth}
        \includegraphics[width=\textwidth]{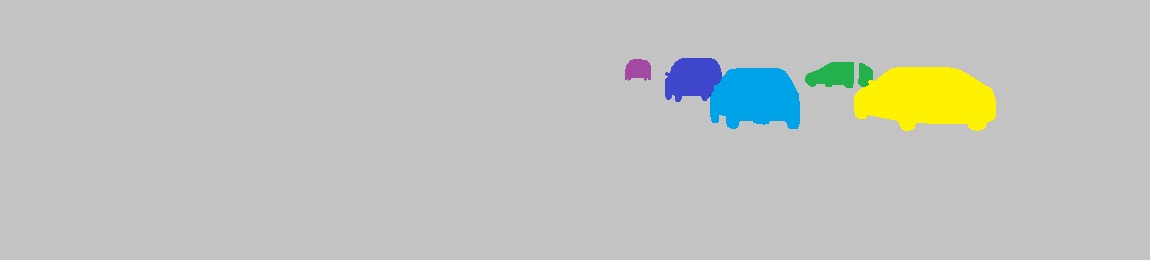}
    \end{subfigure}\hspace{1pt}%
    \begin{subfigure}{0.33\textwidth}
        \includegraphics[width=\textwidth]{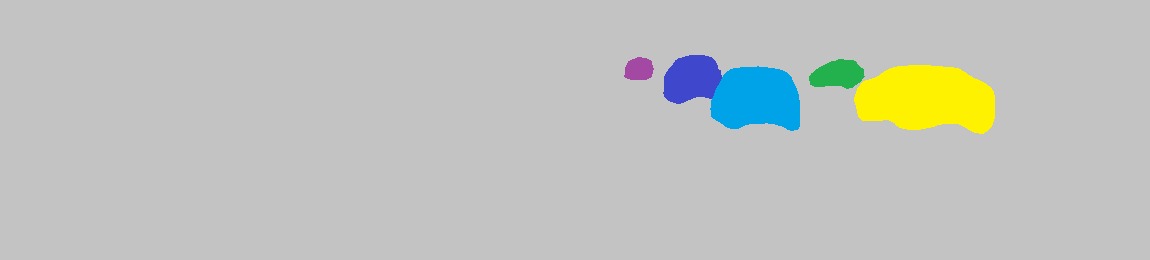}
    \end{subfigure}\\
    \begin{subfigure}{0.33\textwidth}
        \includegraphics[width=\textwidth]{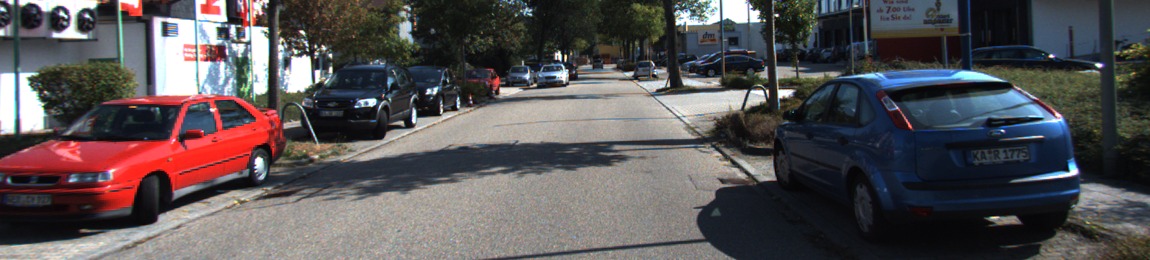}
    \end{subfigure}\hspace{1pt}%
    \begin{subfigure}{0.33\textwidth}
        \includegraphics[width=\textwidth]{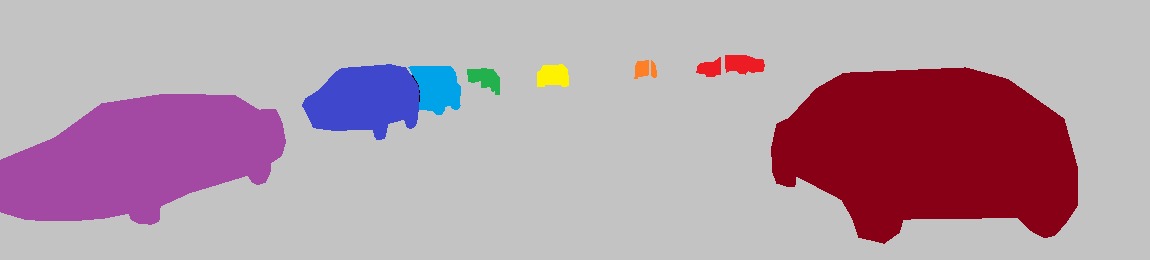}
    \end{subfigure}\hspace{1pt}%
    \begin{subfigure}{0.33\textwidth}
        \includegraphics[width=\textwidth]{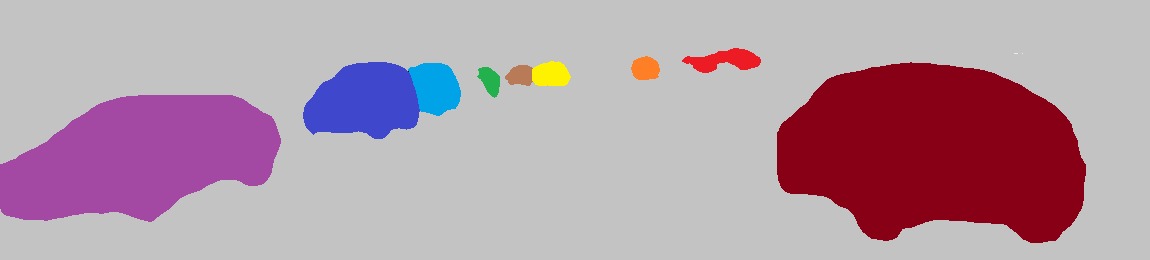}
    \end{subfigure}\\
    \caption{Example results of our instance segmentation (right) and corresponding ground truth (middle) on KITTI.
             We even detect objects at very large distances.\vspace{-0.5em}}
    \label{fig:kittiexamples}
    \end{figure}

    \subsection{Ablation Studies}
    \label{subsec:ablationstudies}

    We evaluated the influence of each proposed component by
    leaving out one or more components from the complete processing pipeline
    (\textit{Ours}). The performance was evaluated on the respective
    validation sets and is listed in \cref{tab:kittiVariantsZhang,tab:csVariants}
    (top) for both datasets.

    For \textit{Ours-D}, we removed the depth channel and chose the template size scale-agnostic.
    It turned out that a rather small template size, which leads to a large number of instance
    proposals, produces the best results. This is possible when post-processing heavily
    relies on correct direction predictions, which induces successful instance fusion.
    However, the performance is significantly worse
    in most metrics on both datasets compared to our full system, which shows that
    the depth information is an essential component of our approach.
    When the fusion component was also removed (\textit{Ours-D-F}), a
    larger template size was needed to prevent an over-segmentation. However, performance dropped
    by an even larger margin than for \textit{Ours-D}. In our last variant
    we kept the depth information but directly used the instance proposals
    as final instance predictions (\textit{Ours-F}). The performance was even slightly worse than
    \textit{Ours-D}, which shows that all our components are important to obtain
    accurate object instances. These observations are consistent on both datasets.

    \subsection{Instance Evaluation}
    \label{subsec:instanceevaluation}

    \myparagraph{KITTI.}
    We clearly outperform all existing works on KITTI~(\textit{Best}~\cite{Urtasun2015a}/\cite{Urtasun2015b}),
    \cf \cref{tab:kittiVariantsZhang} (bottom).
    Compared to the better performing work \textit{Best}~\cite{Urtasun2015b}, we achieve a
    margin of \SI{37}{\percent} relative improvement averaged over all metrics.
    Even when comparing our single variant with the best numbers over all existing variants
    for each metric individually (\textit{Mix}~\cite{Urtasun2015b}),
    we achieve a significantly better performance.
    We also evaluated our approach using the metrics introduced in \cite{Cordts2015} to enable
    comparisons in future publications, \cf \cref{tab:csVariants} (bottom). Qualitative
    results are shown in~\cref{fig:kittiexamples}.

    \setlength{\tabcolsep}{6pt}
    \ctable[
        caption = {Evaluation on Cityscapes \textit{val}
                   (top) and \textit{test} (center)
                   using metrics in \cite{Cordts2015}.
                   Further, we compare the performance for the most
                   frequent label \textit{car}, where we include KITTI \textit{test}
                   (bottom). All numbers are in percent and larger is better.\vspace{-1.2em}},
        label   = {tab:csVariants},
        pos     = {tb},
    ]{lccccccc}{}{\FL
            Variant                     & Dataset    & Labels & $\mapr$   & $\mapr^{50\%}$ & $\mapr^{100\text{m}}$ & $\mapr^{50\text{m}}$ \ML
            Ours-D-F                    & CS val     & all    & $2.4$     & $5.7$          & $3.6$                 & $4.9$                \NN
            Ours-F                      & CS val     & all    & $7.0$     & $17.5$         & $11.1$                & $12.8$               \NN
            Ours-D                      & CS val     & all    & $6.8$     & $15.8$         & $10.9$                & $14.2$               \NN
            Ours                        & CS val     & all    & \bst{9.9} & \bst{22.5}     & \bst{15.3}            & \bst{17.5}           \ML
            MCG+R-CNN \cite{Cordts2015} & CS test    & all    & $4.6$     & $12.9$         & $7.7$                 & $10.3$               \NN
            Ours                        & CS test    & all    & \bst{8.9} & \bst{21.1}     & \bst{15.3}            & \bst{16.7}           \ML
            MCG+R-CNN \cite{Cordts2015} & CS test    & car    & $10.5$    & $26.0$         & $17.5$                & $21.2$               \NN
            Ours                        & CS test    & car    & $22.5$    & $37.8$         & $36.4$                & $40.7$               \NN
            Ours                        & KITTI test & car    & $41.6$    & $69.1$         & $49.3$                & $49.3$               \LL
    }

    \begin{figure}[t]
    \centering
    \begin{subfigure}{0.33\textwidth}
        \caption{Input Image}
        \includegraphics[width=\textwidth]{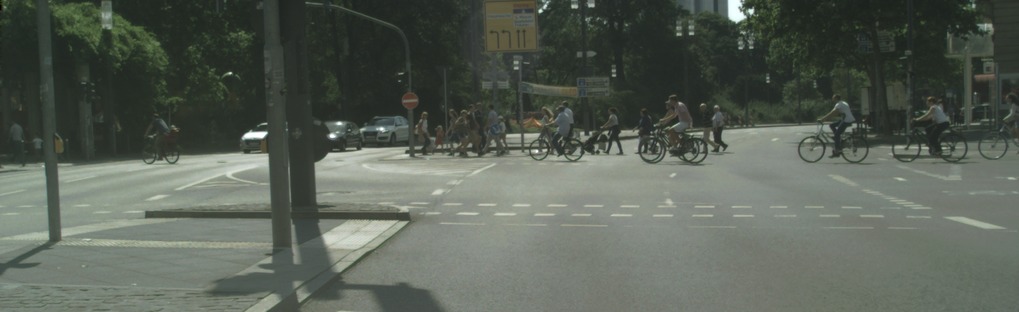}
    \end{subfigure}\hspace{1pt}%
    \begin{subfigure}{0.33\textwidth}
        \caption{Instance Ground Truth}
        \includegraphics[width=\textwidth]{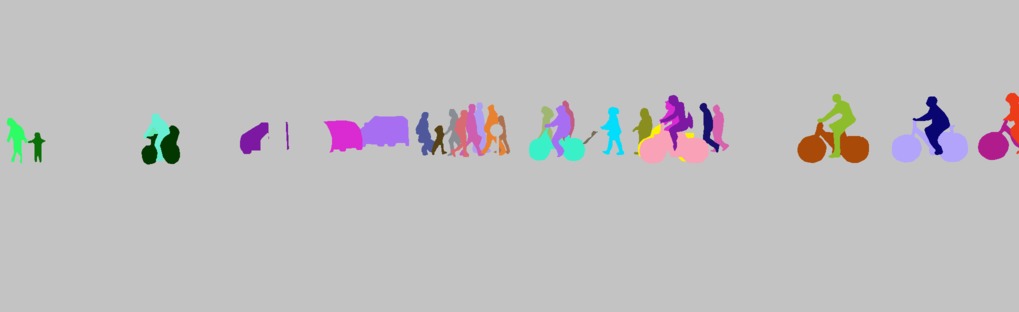}
    \end{subfigure}\hspace{1pt}%
    \begin{subfigure}{0.33\textwidth}
        \caption{Instance Prediction}
        \includegraphics[width=\textwidth]{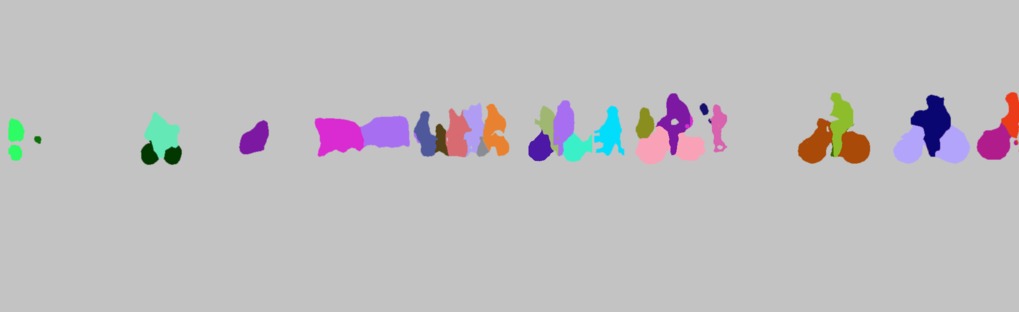}
    \end{subfigure}\\
    \begin{subfigure}{0.33\textwidth}
        \includegraphics[width=\textwidth]{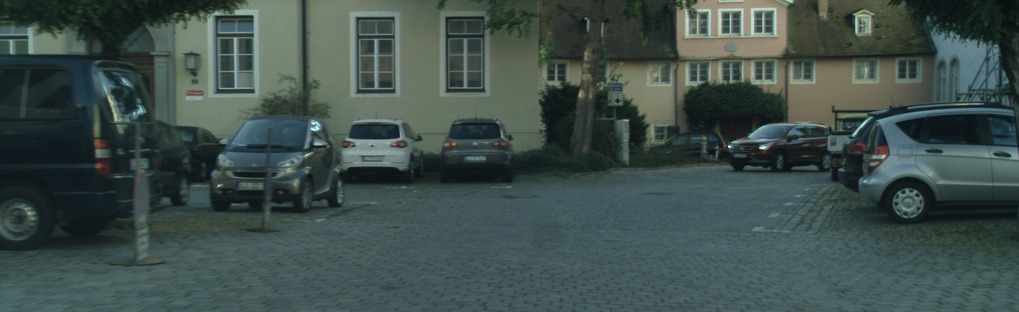}
    \end{subfigure}\hspace{1pt}%
    \begin{subfigure}{0.33\textwidth}
        \includegraphics[width=\textwidth]{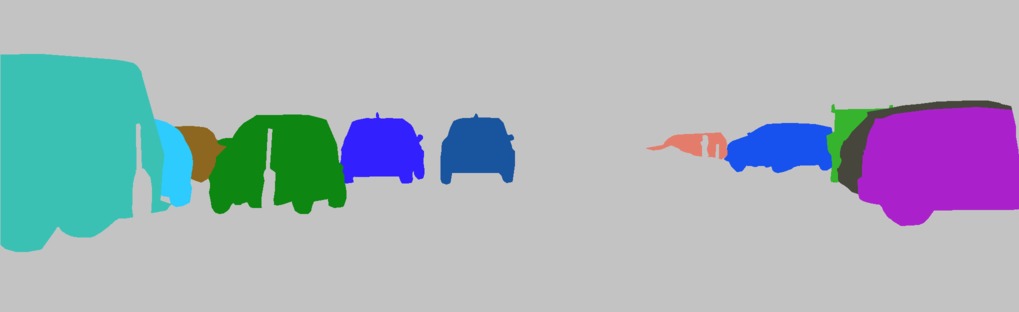}
    \end{subfigure}\hspace{1pt}%
    \begin{subfigure}{0.33\textwidth}
        \includegraphics[width=\textwidth]{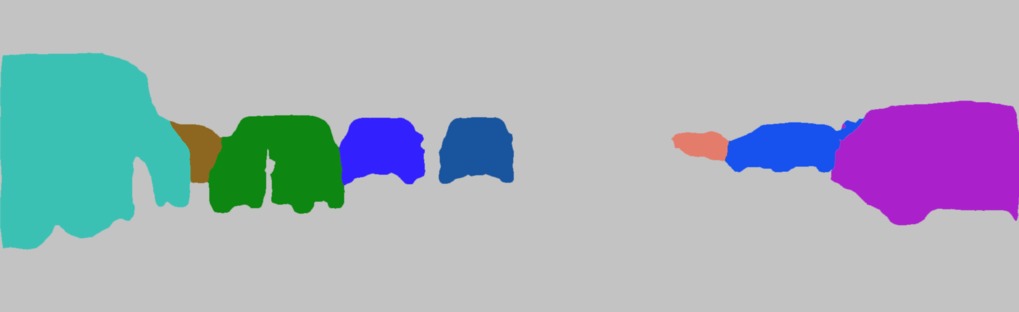}
    \end{subfigure}\\
    \begin{subfigure}{0.33\textwidth}
        \includegraphics[width=\textwidth]{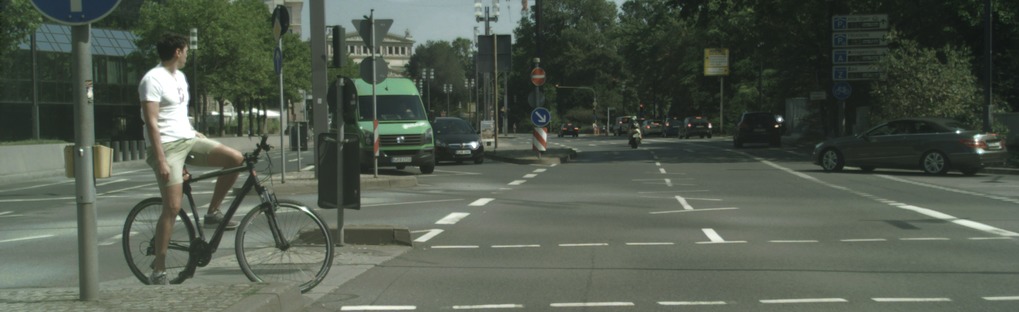}
    \end{subfigure}\hspace{1pt}%
    \begin{subfigure}{0.33\textwidth}
        \includegraphics[width=\textwidth]{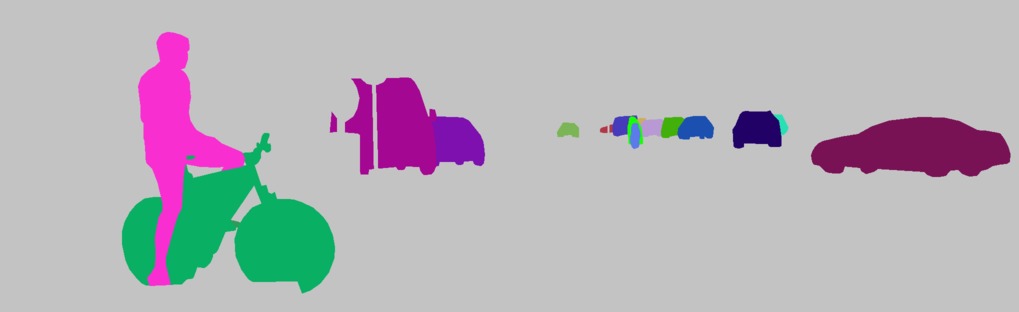}
    \end{subfigure}\hspace{1pt}%
    \begin{subfigure}{0.33\textwidth}
        \includegraphics[width=\textwidth]{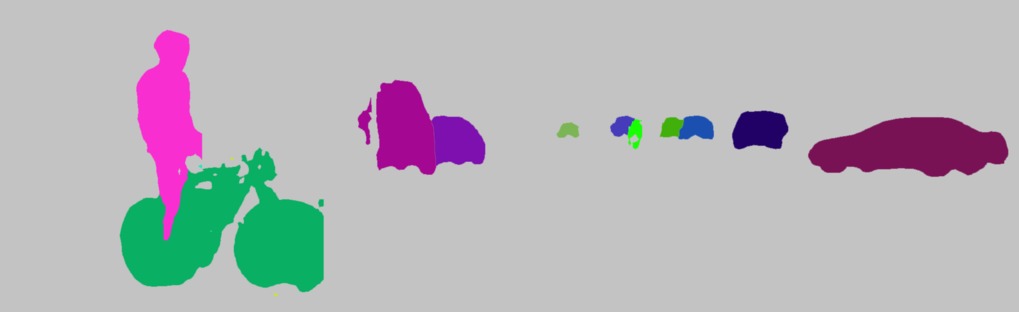}
    \end{subfigure}\\
    \vspace{5pt}
    \begin{subfigure}{0.33\textwidth}
        \includegraphics[width=\textwidth]{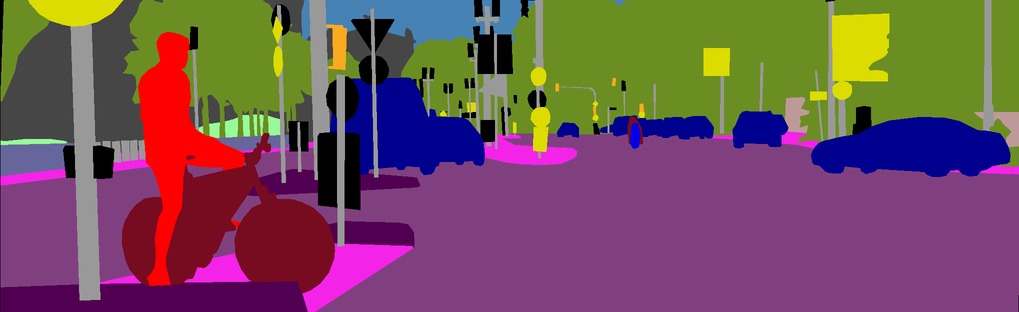}
    \end{subfigure}\hspace{1pt}%
    \begin{subfigure}{0.33\textwidth}
        \includegraphics[width=\textwidth]{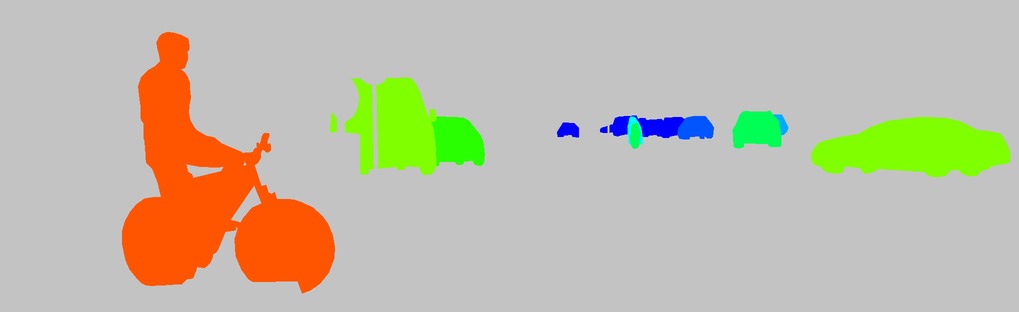}
    \end{subfigure}\hspace{1pt}%
    \begin{subfigure}{0.33\textwidth}
        \includegraphics[width=\textwidth]{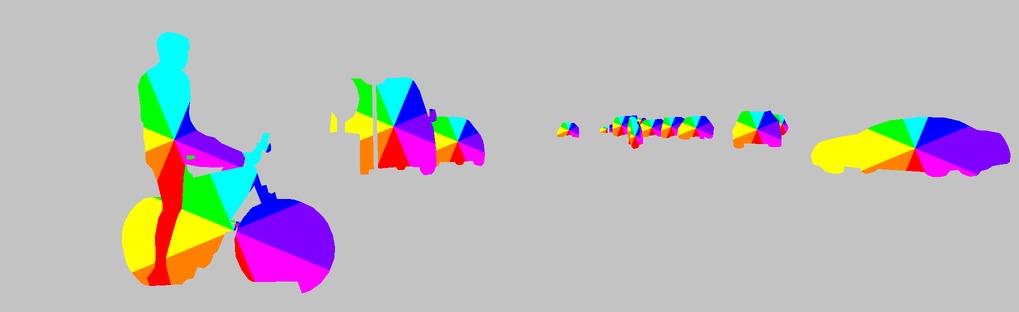}
    \end{subfigure}\\
    \begin{subfigure}{0.33\textwidth}
        \includegraphics[width=\textwidth]{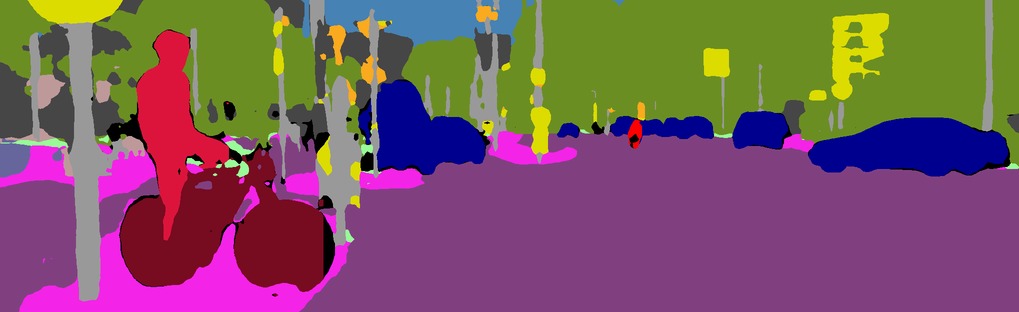}
        \caption{Semantics}
    \end{subfigure}\hspace{1pt}%
    \begin{subfigure}{0.33\textwidth}
        \includegraphics[width=\textwidth]{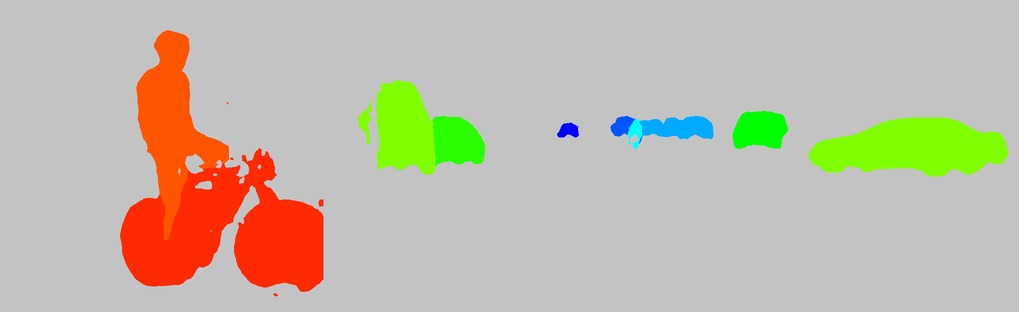}
        \caption{Depth}
    \end{subfigure}\hspace{1pt}%
    \begin{subfigure}{0.33\textwidth}
        \includegraphics[width=\textwidth]{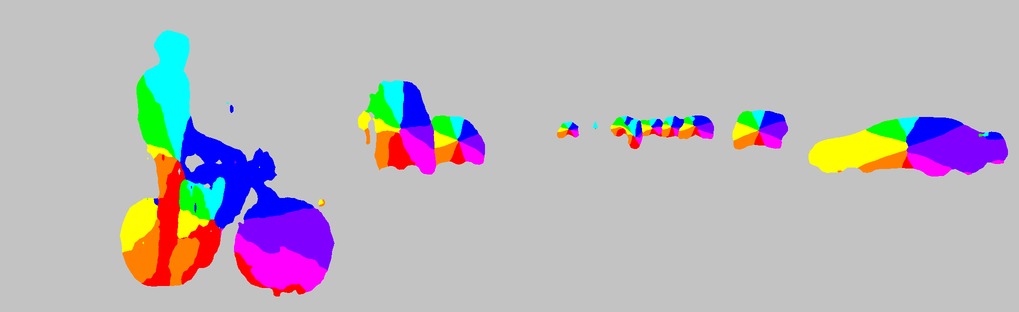}
        \caption{Direction}
    \end{subfigure}\\
    \caption{Example results of our instance segmentation and corresponding
                ground truth (rows $1$--$3$) on Cityscapes. We also include the
                three FCN output channels (row $5$) and their ground truth (row $4$).
                It can be seen that even distant objects are segmented well and
                the approach can handle occlusions.\vspace{-0.5em}}
    \label{fig:cityscapesexamples}
    \end{figure}

    \myparagraph{Cityscapes.}
    On the Cityscapes dataset, our approach outperforms the baseline
    \textit{MCG+R-CNN}~\cite{Cordts2015} in all proposed metrics as evaluated by the
    dataset's submission server, \cf \cref{tab:csVariants} (center).
    We nearly double the performance in terms of the main score $\mapr$.
    Compared to the performance on KITTI, \cf \cref{tab:csVariants} (bottom), the numbers
    are significantly lower, indicating the higher complexity of scenes in Cityscapes.
    Qualitative results are shown in~\cref{fig:cityscapesexamples}.

    \setlength{\tabcolsep}{6pt}
    \ctable[
        caption = {Instance-based depth evaluation on KITTI test
                   and Cityscapes validation. MAE and RMSE are in meters, the others
                   in percent. MAE, RMSE, and ARD denote error metrics, where smaller
                   is better, $\delta_i$ represent accuracy, where higher is better.},
        label   = {tab:depthEval},
        pos     = {t},
        width   = .85\textwidth
    ]{Xcccccc}{}{           \FL
            Dataset          & MAE   & RMSE   & ARD    & $\delta_1$ & $\delta_2$ & $\delta_3$  \ML
            KITTI (test)     & $1.7$ & $2.8$  & $7.7$  & $95.1$     & $99.3$     & $99.8$      \NN
            Cityscapes (val) & $7.7$ & $24.8$ & $11.3$ & $86.2$     & $95.1$     & $97.7$      \LL
    }

    \subsection{Depth Evaluation}
    As shown in \cref{tab:depthEval}, the average relative and
    mean absolute error of
    our predicted instances are as low as \SI{7.7}{\percent} and
    \SI{1.7}{\metre}, respectively, on the KITTI dataset.
    On the Cityscapes dataset, which contains much more complex scenes,
    with many and distant object instances, we achieve
    \SI{11.3}{\percent} and \SI{7.7}{\metre}, respectively.
    These results are particularly impressive, since we used
    only single monocular images as input for our network. We hope
    that future publications compare their depth estimation
    performance using the proposed metrics.

    \setlength{\tabcolsep}{6pt}
    \ctable[
        caption = {Semantic pixel-level evaluation on Cityscapes test compared to baselines and
                   using the corresponding metrics~\cite{Cordts2015}. All values are
                   in percent and larger is better.},
        label   = {tab:semanticEval},
        pos     = {t},
        width   = .85\textwidth
    ]{Xcccc}{}{\FL
            Method                       &
            $\miou_{\finelevel}$         &
            $\miiou_{\finelevel}$        &
            $\miou_{\coarselevel}$       &
            $\miiou_{\coarselevel}$      \ML
            FCN 8s \cite{Cordts2015} & $65.3$     & $41.7$     & $85.7$     & $70.1$     \NN
            Dilation10 \cite{Yu2016} & \bst{67.1} & \bst{42.0} & \bst{86.5} & $71.1$     \NN
            Ours                     & $64.3$     & $41.6$     & $85.9$     & \bst{73.9} \LL
    }

    \subsection{Evaluation of semantic class labels}
    Our method also yields a pixel-level semantic
    labeling including background classes that we evaluate on Cityscapes, \cf
    \cref{tab:semanticEval}. We compare to two baselines, \textit{FCN
    8s}~\cite{FCN2015} that uses the same FCN architecture as our approach and
    \textit{Dilation10}~\cite{Yu2016}, which is the currently best
    performing approach on Cityscapes~\cite{Cordts2015}. It can be seen
    that our approach is on par with the state-of-the-art although this
    work focuses on the harder instance segmentation task.

    \section{Conclusion}
    \label{sec:conclusion}
    In this work, we present a fully convolutional network that
    predicts pixel-wise depth, semantics, and instance-level direction
    cues to reach an excellent level of holistic scene understanding. Instead of complex
    architectures or graphical models for post-processing, our approach
    performs well using only standard computer vision techniques
    applied to the network's three output channels. Our approach
    does not depend on region proposals and scales well
    for arbitrary numbers of object instances in an image.

    We outperform existing works on the challenging urban street scene datasets
    Cityscapes~\cite{Cordts2015} and KITTI~\cite{Urtasun2015a,Urtasun2015b} by a
    large margin. On KITTI, our approach achieves \SI{37}{\percent} relative
    improvement averaged over all metrics and we almost double the performance on
    Cityscapes. As our approach can reliably predict absolute depth values per instance,
    we provide an instance-based depth evaluation. Our depth predictions achieve a relative error of only a
    few meters, even though the datasets contain instances in more than one hundred meters
    distance. The main focus of this work is instance segmentation, but we also
    achieve state-of-the-art performance for pixel-level semantic labeling
    on Cityscapes, with a new best performance on an instance-based score
    over categories.

    \bibliographystyle{splncs03}
    \bibliography{bib}

    \appendix
    \pagestyle{headings}
    \mainmatter
    \pagenumbering{roman}
    \title{Supplementary Material for\\
            Pixel-level Encoding and Depth Layering for Instance-level Semantic Labeling}
    \author{Jonas Uhrig$^{1,2}$, Marius Cordts$^{1,3}$, Uwe Franke$^{1}$, Thomas Brox$^{2}$}
    \institute{$^1$Daimler AG R\&D, $^2$University of Freiburg, $^3$TU Darmstadt\\\texttt{jonas.uhrig@daimler.com}}
    \titlerunning{Sup. Mat.: Pixel Enc. and Depth Layering for Instance-level Sem. Labeling}
    \authorrunning{ Jonas Uhrig, Marius Cordts, Uwe Franke, Thomas Brox }
    \maketitle

    \section{Qualitative Results}
\begin{figure}[p]
    \centering 
    \begin{subfigure}{0.33\textwidth}
        \includegraphics[width=\textwidth]{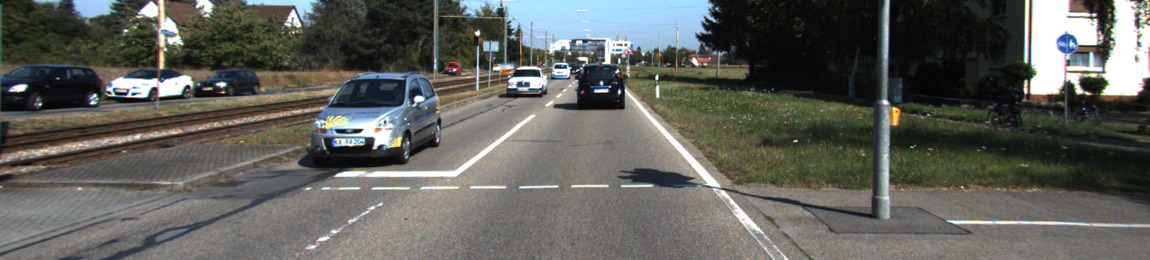}
    \end{subfigure}\hspace{1pt}%
    \begin{subfigure}{0.33\textwidth}
        \includegraphics[width=\textwidth]{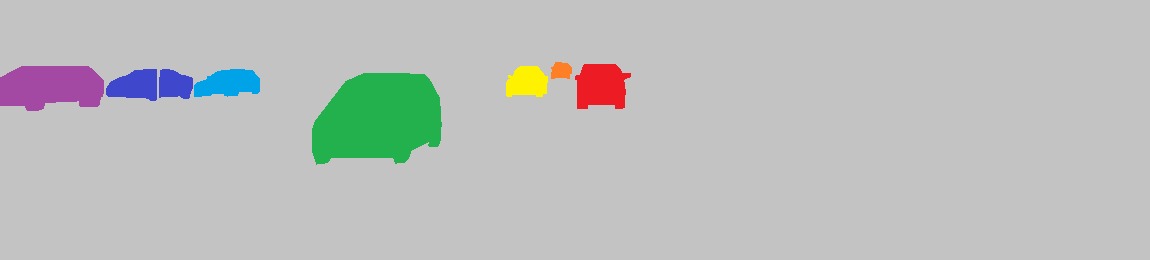}
    \end{subfigure}\hspace{1pt}%
    \begin{subfigure}{0.33\textwidth}
        \includegraphics[width=\textwidth]{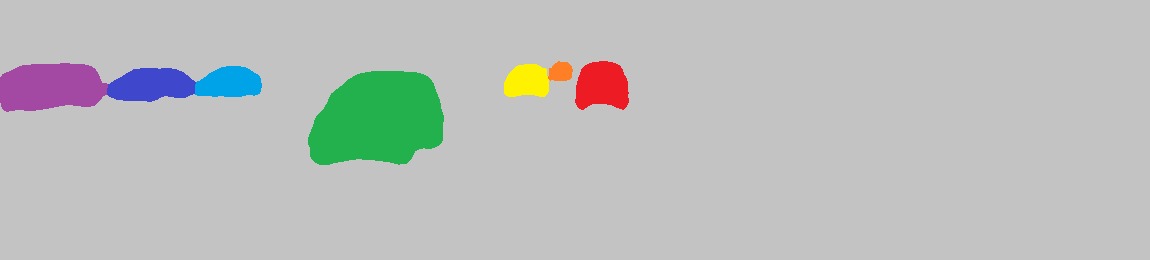}
    \end{subfigure}\\
    \begin{subfigure}{0.33\textwidth}
        \includegraphics[width=\textwidth]{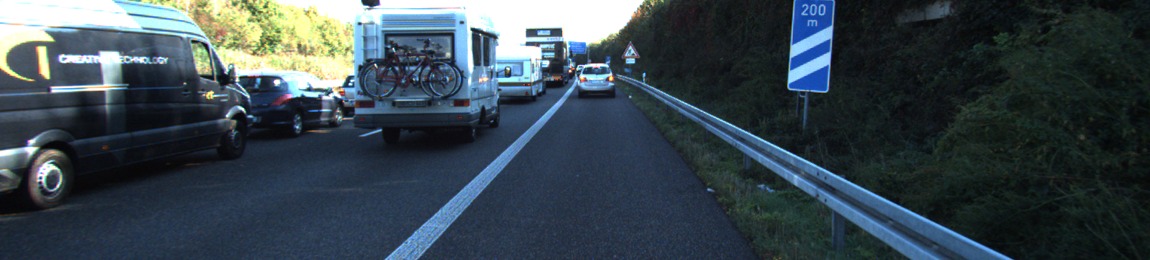}
    \end{subfigure}\hspace{1pt}%
    \begin{subfigure}{0.33\textwidth}
        \includegraphics[width=\textwidth]{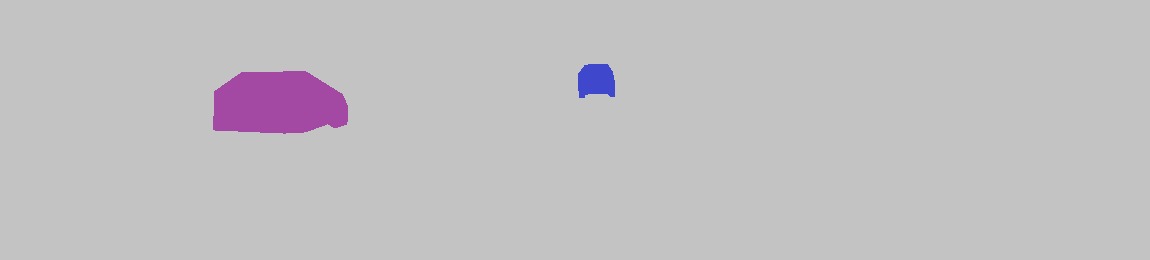}
    \end{subfigure}\hspace{1pt}%
    \begin{subfigure}{0.33\textwidth}
        \includegraphics[width=\textwidth]{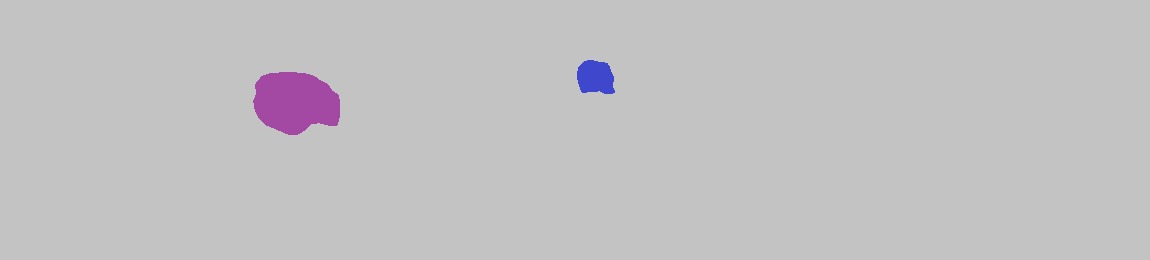}
    \end{subfigure}\\
    \begin{subfigure}{0.33\textwidth}
        \includegraphics[width=\textwidth]{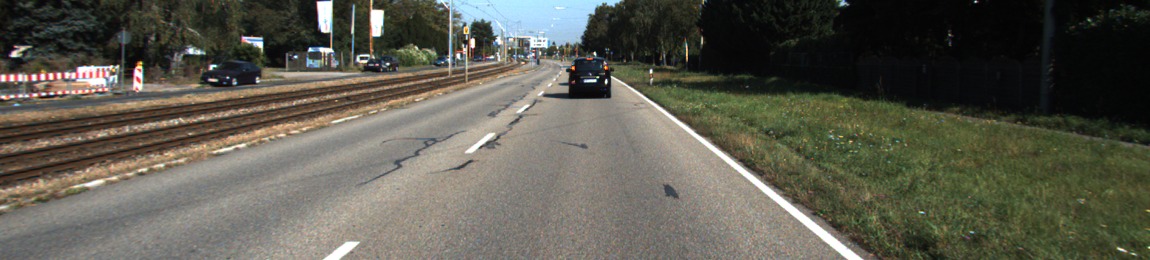}
    \end{subfigure}\hspace{1pt}%
    \begin{subfigure}{0.33\textwidth}
        \includegraphics[width=\textwidth]{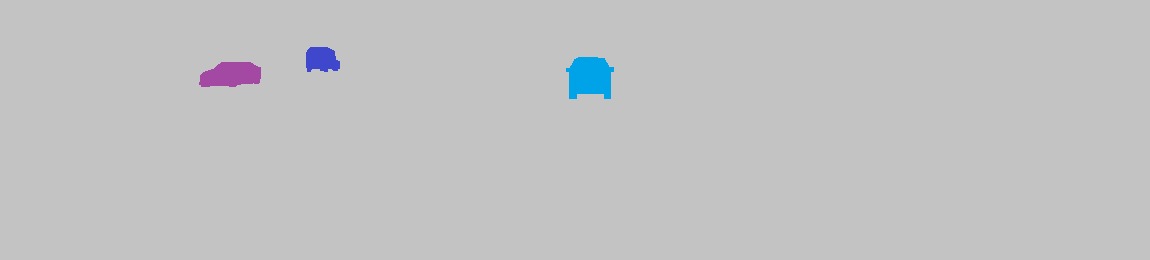}
    \end{subfigure}\hspace{1pt}%
    \begin{subfigure}{0.33\textwidth}
        \includegraphics[width=\textwidth]{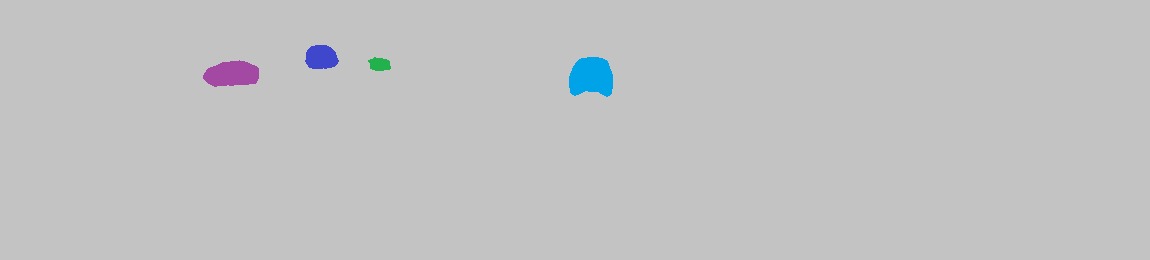}
    \end{subfigure}\\
    \begin{subfigure}{0.33\textwidth}
        \includegraphics[width=\textwidth]{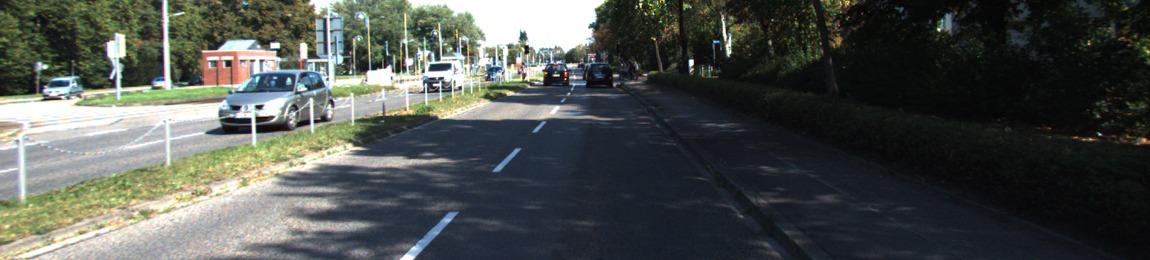}
    \end{subfigure}\hspace{1pt}%
    \begin{subfigure}{0.33\textwidth}
        \includegraphics[width=\textwidth]{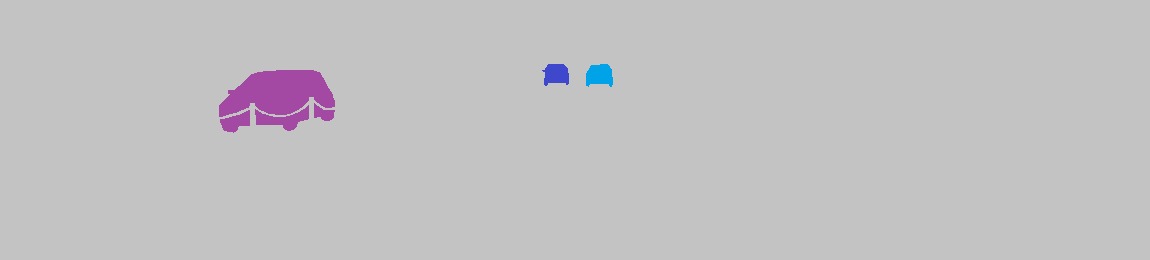}
    \end{subfigure}\hspace{1pt}%
    \begin{subfigure}{0.33\textwidth}
        \includegraphics[width=\textwidth]{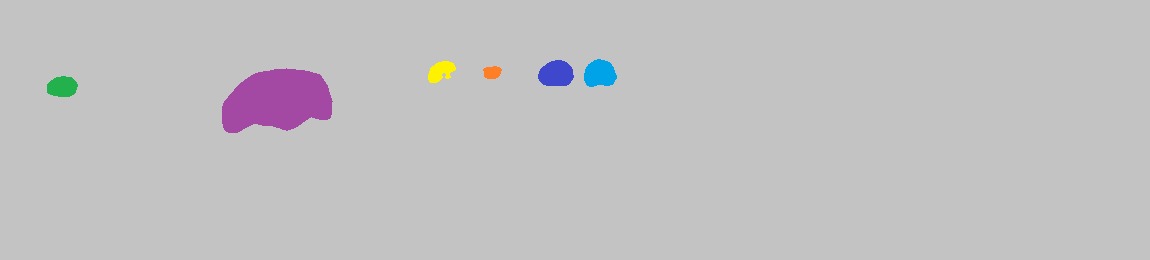}
    \end{subfigure}\\
    \begin{subfigure}{0.33\textwidth}
        \includegraphics[width=\textwidth]{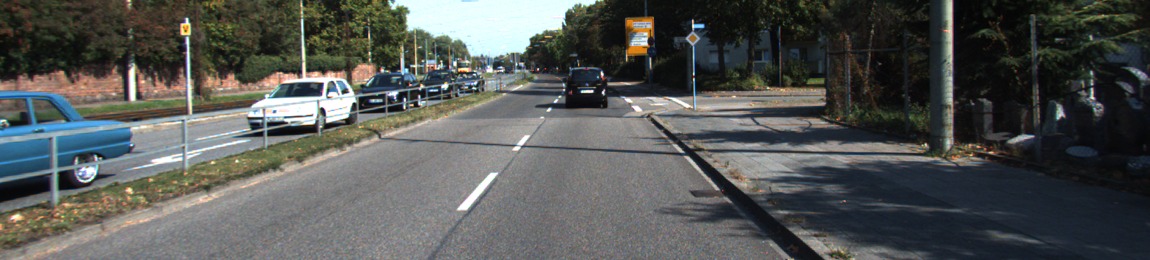}
    \end{subfigure}\hspace{1pt}%
    \begin{subfigure}{0.33\textwidth}
        \includegraphics[width=\textwidth]{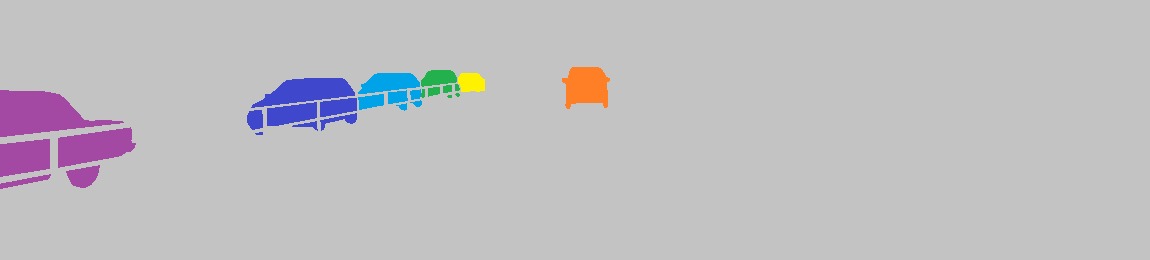}
    \end{subfigure}\hspace{1pt}%
    \begin{subfigure}{0.33\textwidth}
        \includegraphics[width=\textwidth]{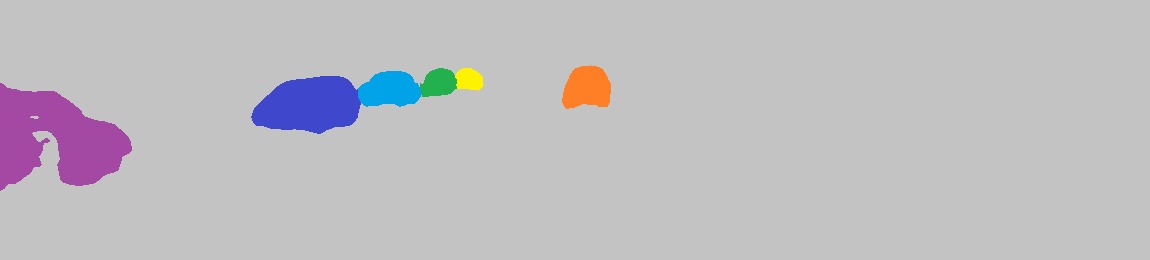}
    \end{subfigure}\\
    \begin{subfigure}{0.33\textwidth}
        \includegraphics[width=\textwidth]{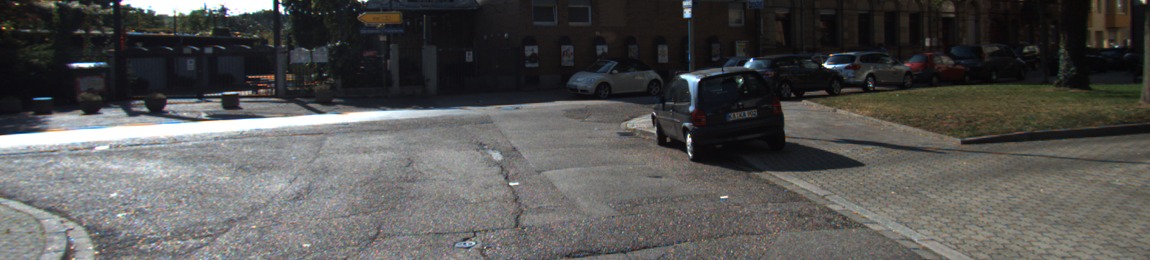}
    \end{subfigure}\hspace{1pt}%
    \begin{subfigure}{0.33\textwidth}
        \includegraphics[width=\textwidth]{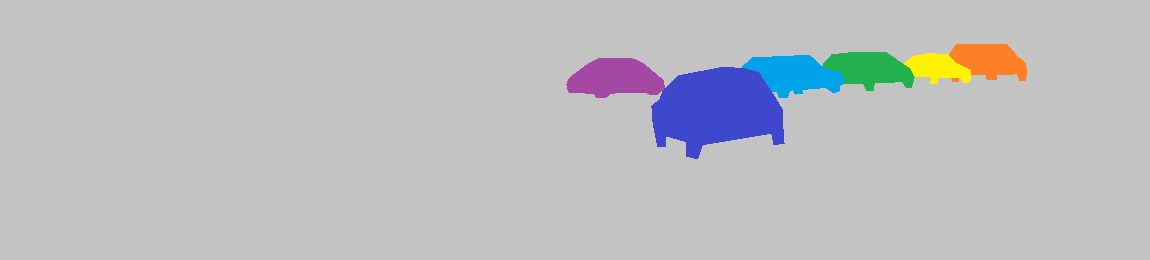}
    \end{subfigure}\hspace{1pt}%
    \begin{subfigure}{0.33\textwidth}
        \includegraphics[width=\textwidth]{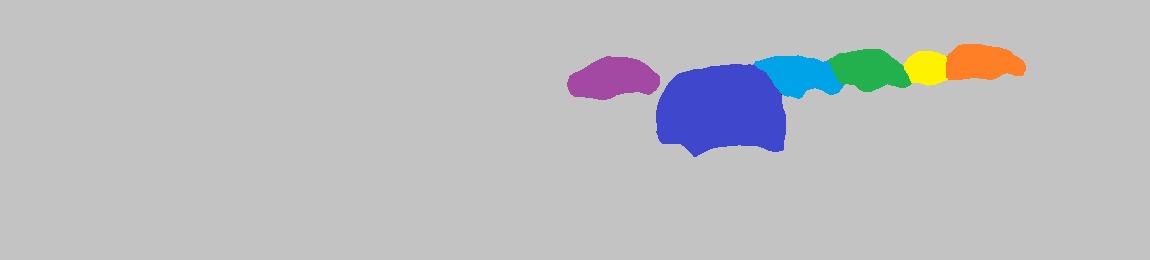}
    \end{subfigure}\\
    \begin{subfigure}{0.33\textwidth}
        \includegraphics[width=\textwidth]{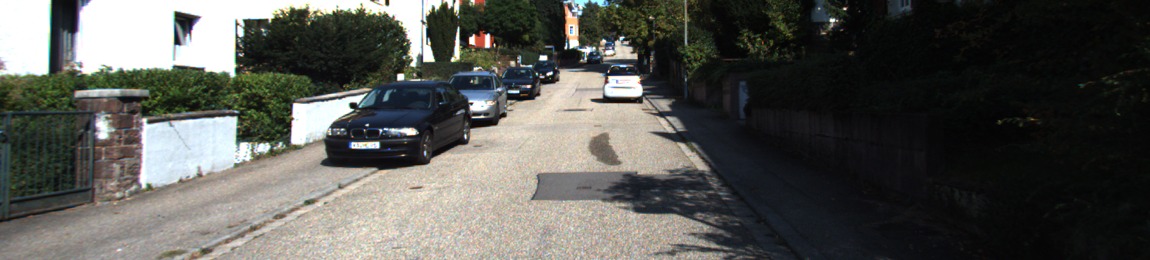}
    \end{subfigure}\hspace{1pt}%
    \begin{subfigure}{0.33\textwidth}
        \includegraphics[width=\textwidth]{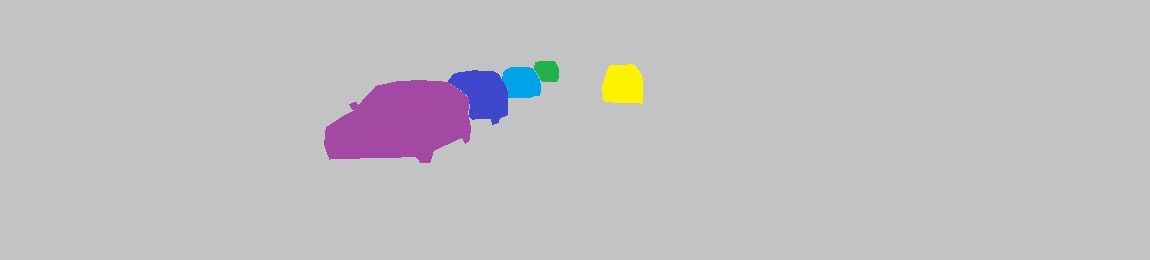}
    \end{subfigure}\hspace{1pt}%
    \begin{subfigure}{0.33\textwidth}
        \includegraphics[width=\textwidth]{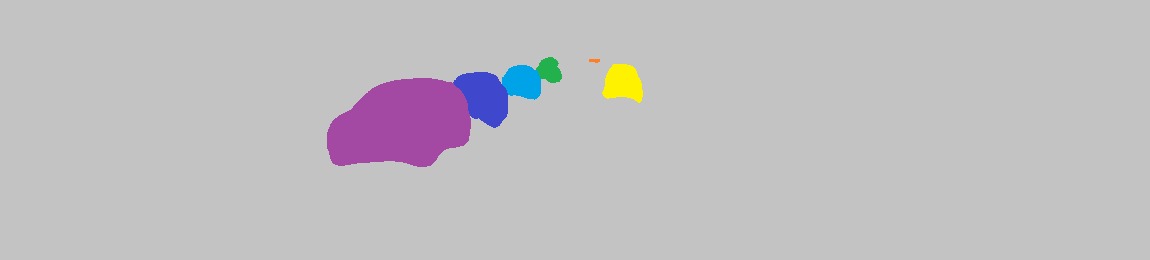}
    \end{subfigure}\\
    \begin{subfigure}{0.33\textwidth}
        \includegraphics[width=\textwidth]{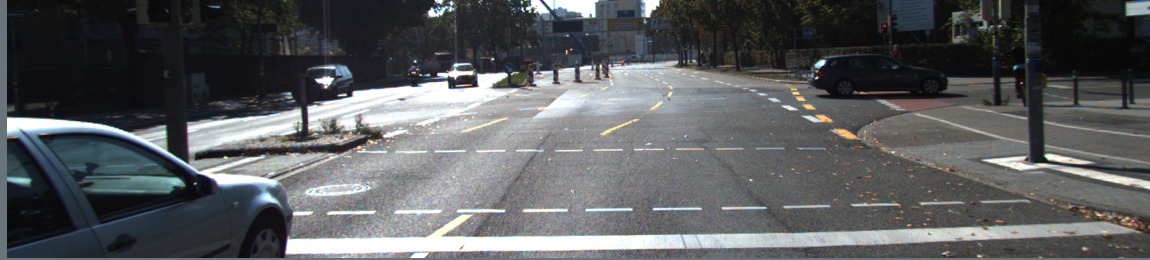}
        \caption{Input Image}
    \end{subfigure}\hspace{1pt}%
    \begin{subfigure}{0.33\textwidth}
        \includegraphics[width=\textwidth]{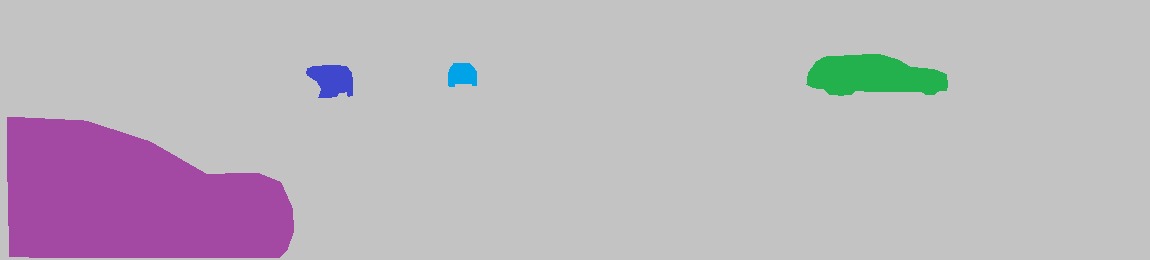}
        \caption{Instance Ground Truth}
    \end{subfigure}\hspace{1pt}%
    \begin{subfigure}{0.33\textwidth}
        \includegraphics[width=\textwidth]{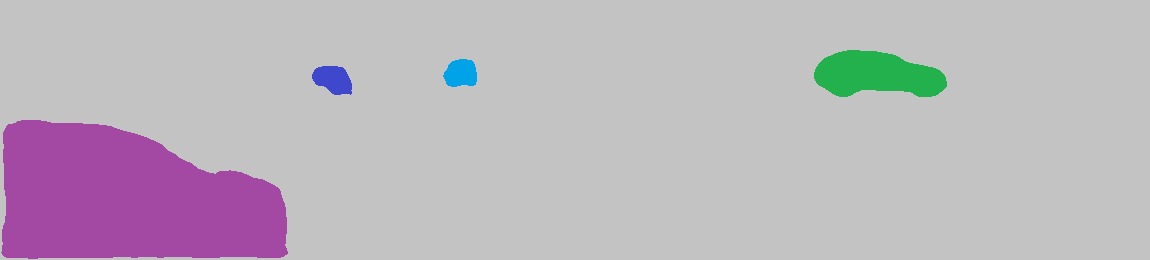}
        \caption{Instance Prediction}
    \end{subfigure}\\
    \caption{Further example results of our instance segmentation (right) and corresponding
                     ground truth (middle) on KITTI.\vspace{-0.5em}}
    \label{fig:kittiadditionalexamples}
    \end{figure}

\begin{figure}[p]
    \centering 
    \begin{subfigure}{0.33\textwidth}
        \includegraphics[width=\textwidth]{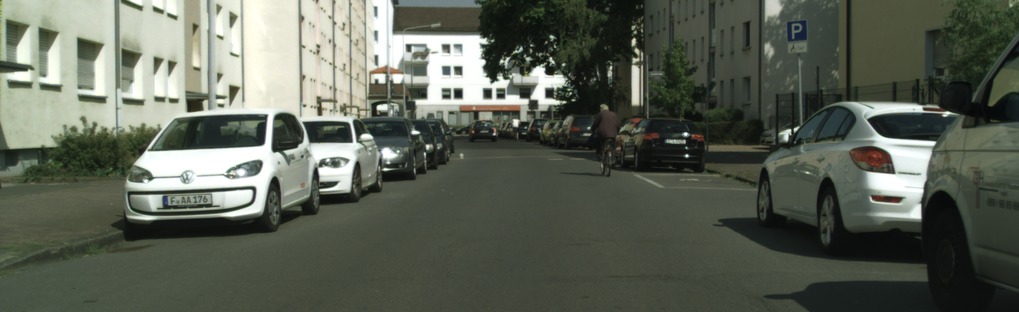}
    \end{subfigure}\hspace{1pt}%
    \begin{subfigure}{0.33\textwidth}
        \includegraphics[width=\textwidth]{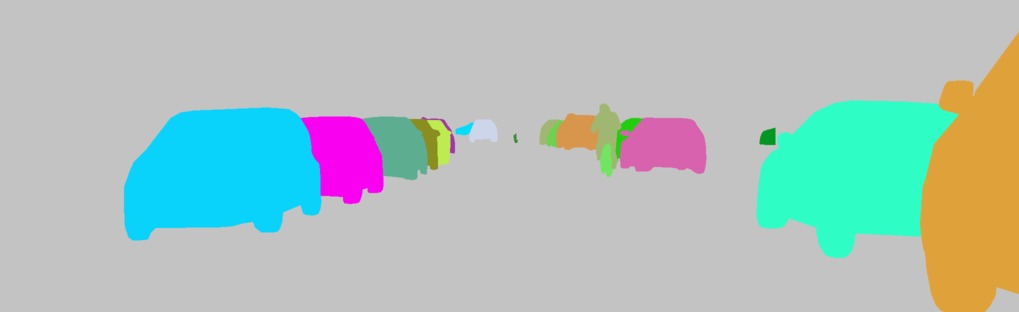}
    \end{subfigure}\hspace{1pt}%
    \begin{subfigure}{0.33\textwidth}
        \includegraphics[width=\textwidth]{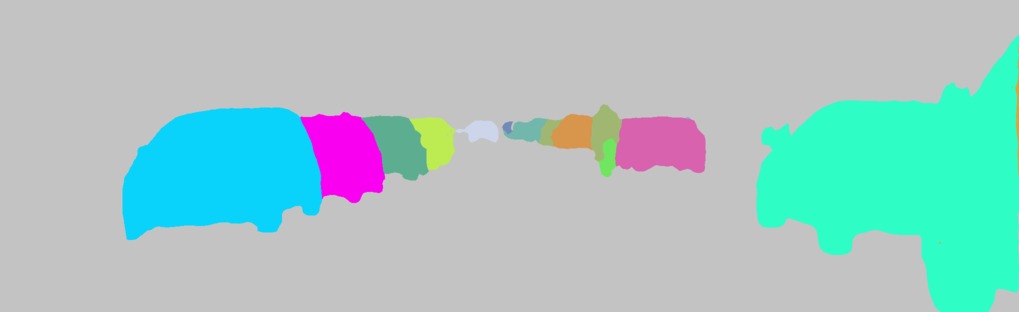}
    \end{subfigure}\\
    \begin{subfigure}{0.33\textwidth}
        \includegraphics[width=\textwidth]{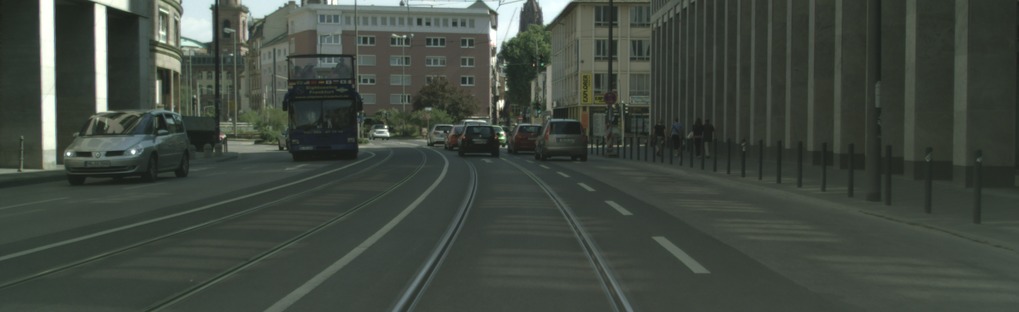}
    \end{subfigure}\hspace{1pt}%
    \begin{subfigure}{0.33\textwidth}
        \includegraphics[width=\textwidth]{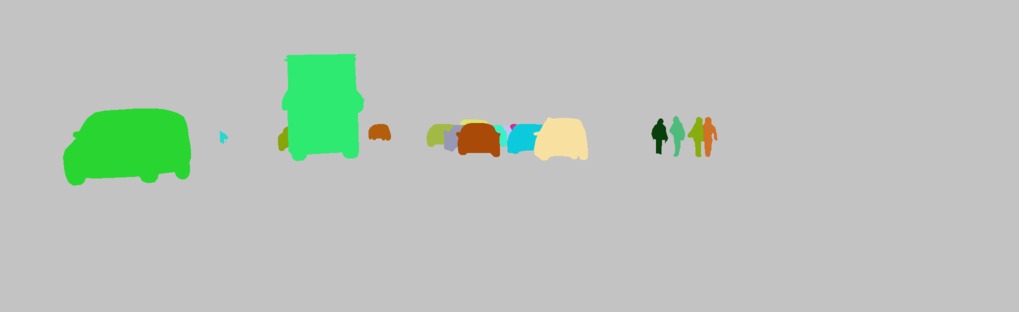}
    \end{subfigure}\hspace{1pt}%
    \begin{subfigure}{0.33\textwidth}
        \includegraphics[width=\textwidth]{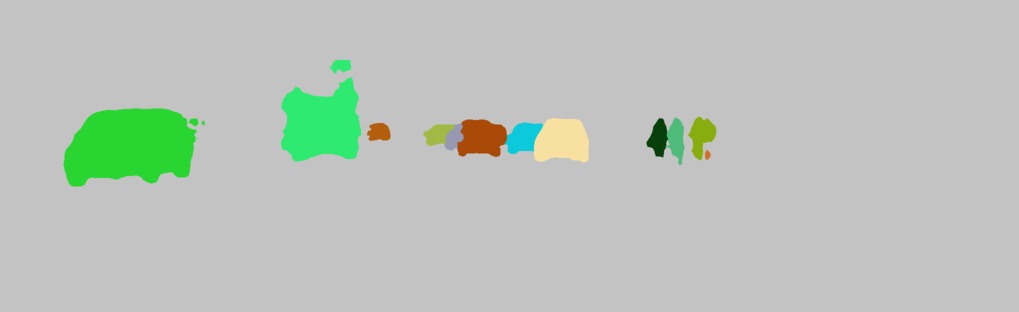}
    \end{subfigure}\\
    \begin{subfigure}{0.33\textwidth}
        \includegraphics[width=\textwidth]{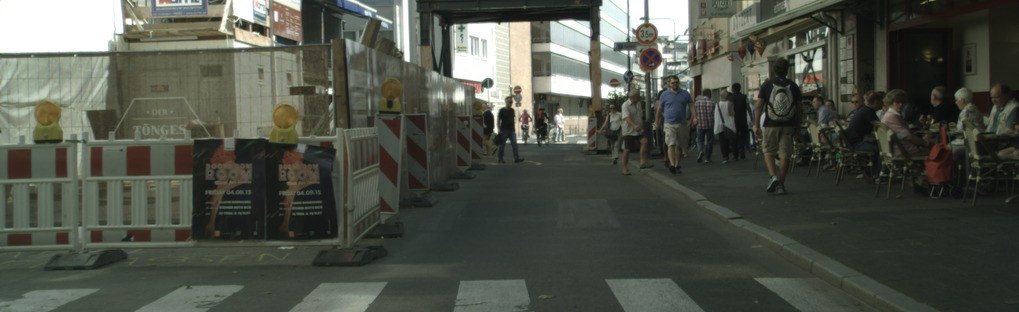}
    \end{subfigure}\hspace{1pt}%
    \begin{subfigure}{0.33\textwidth}
        \includegraphics[width=\textwidth]{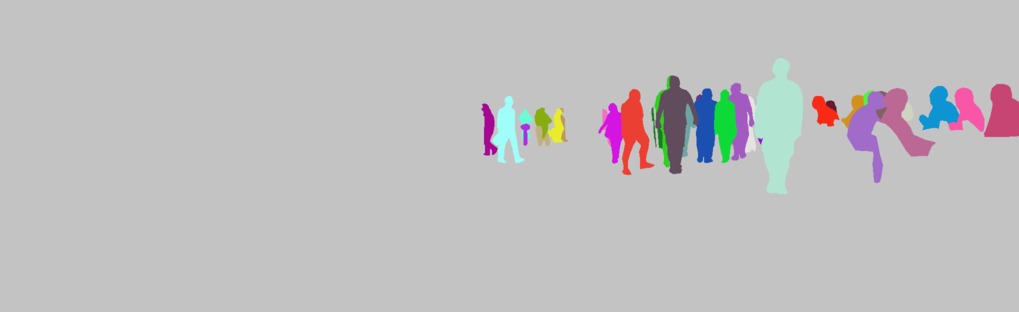}
    \end{subfigure}\hspace{1pt}%
    \begin{subfigure}{0.33\textwidth}
        \includegraphics[width=\textwidth]{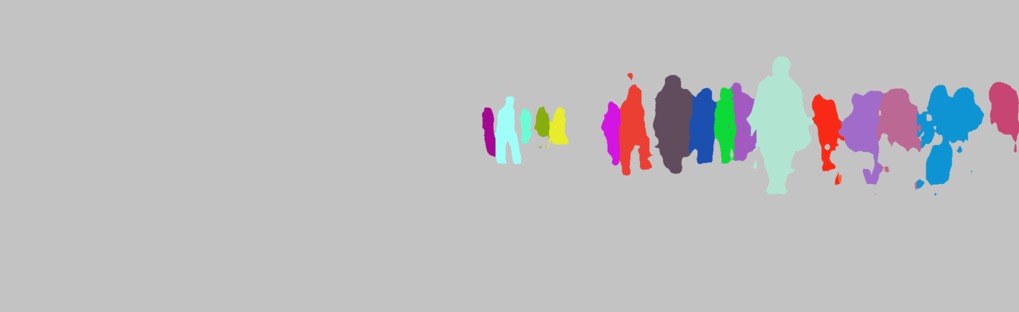}
    \end{subfigure}\\
    \begin{subfigure}{0.33\textwidth}
        \includegraphics[width=\textwidth]{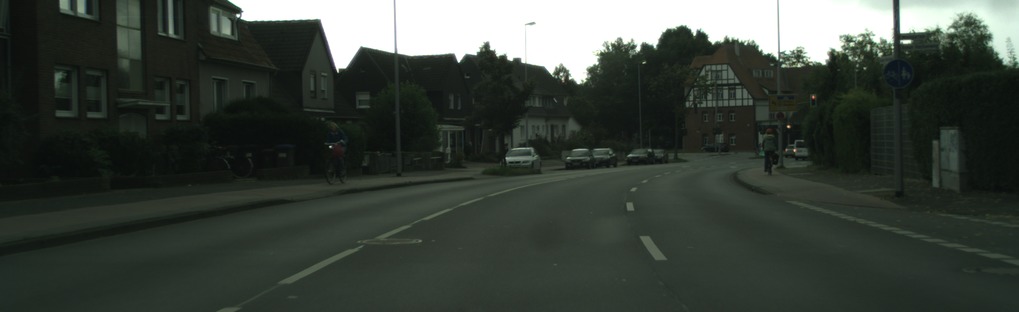}
    \end{subfigure}\hspace{1pt}%
    \begin{subfigure}{0.33\textwidth}
        \includegraphics[width=\textwidth]{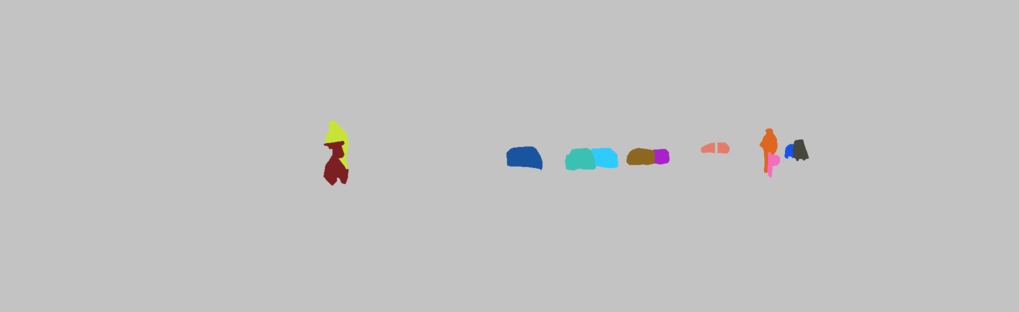}
    \end{subfigure}\hspace{1pt}%
    \begin{subfigure}{0.33\textwidth}
        \includegraphics[width=\textwidth]{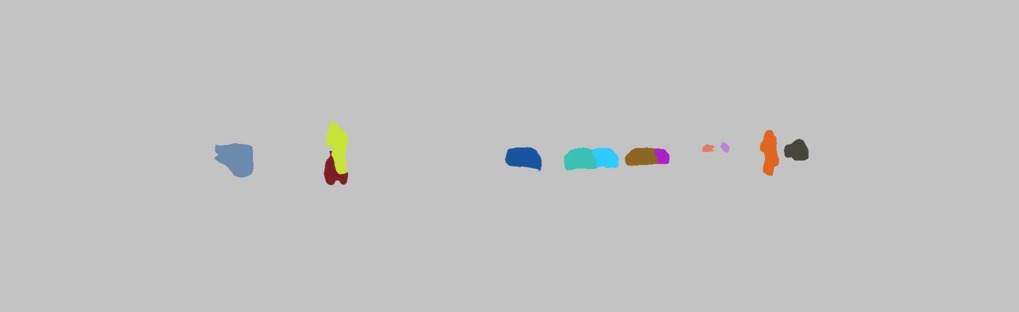}
    \end{subfigure}\\
    \begin{subfigure}{0.33\textwidth}
        \includegraphics[width=\textwidth]{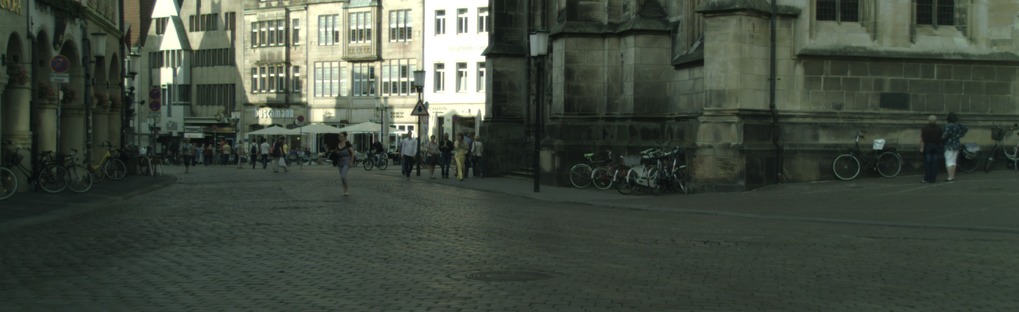}
    \end{subfigure}\hspace{1pt}%
    \begin{subfigure}{0.33\textwidth}
        \includegraphics[width=\textwidth]{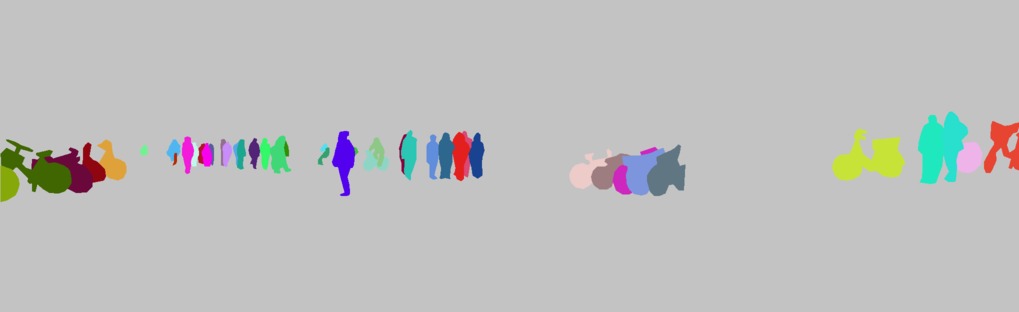}
    \end{subfigure}\hspace{1pt}%
    \begin{subfigure}{0.33\textwidth}
        \includegraphics[width=\textwidth]{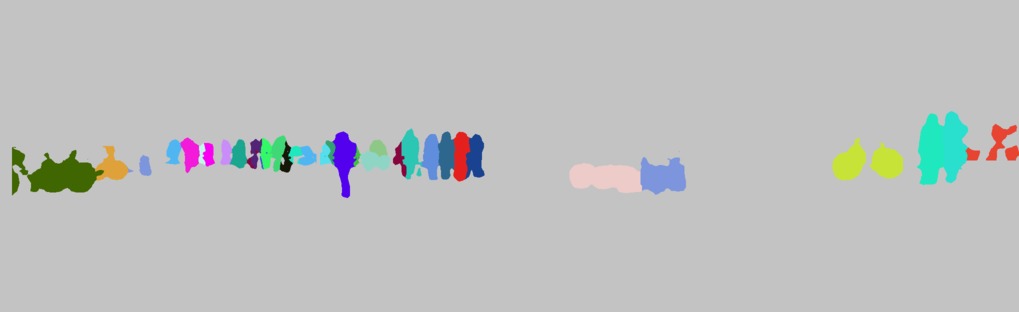}
    \end{subfigure}\\
    \begin{subfigure}{0.33\textwidth}
        \includegraphics[width=\textwidth]{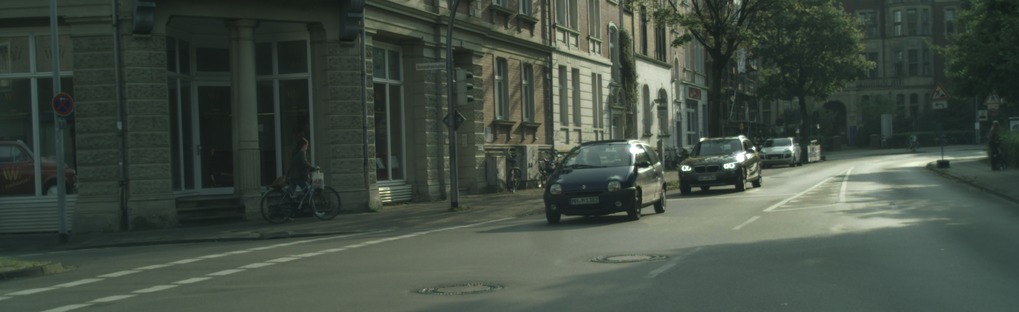}
        \caption{Input Image}
    \end{subfigure}\hspace{1pt}%
    \begin{subfigure}{0.33\textwidth}
        \includegraphics[width=\textwidth]{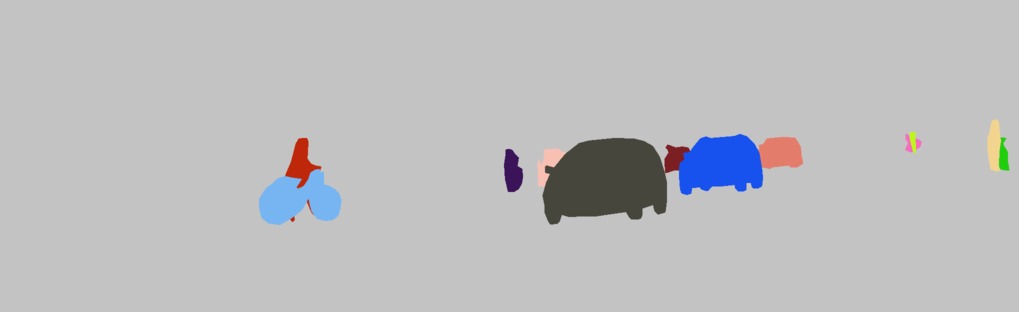}
        \caption{Instance Ground Truth}
    \end{subfigure}\hspace{1pt}%
    \begin{subfigure}{0.33\textwidth}
        \includegraphics[width=\textwidth]{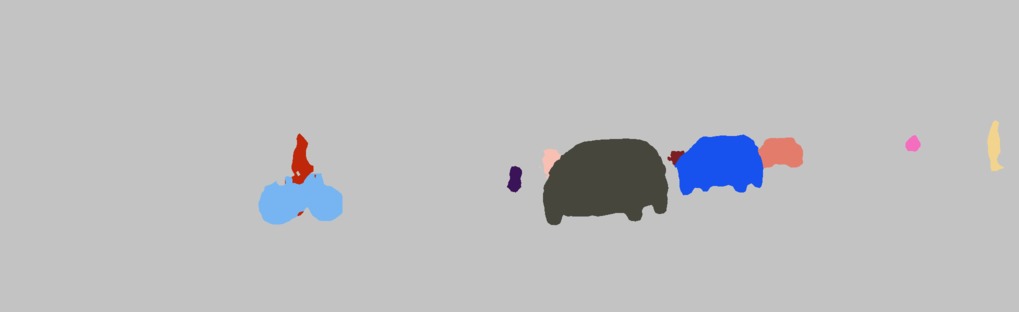}
        \caption{Instance Prediction}
    \end{subfigure}\\
    \caption{Further example results of our instance segmentation (right) and corresponding
                ground truth (center) on Cityscapes \textit{validation}.}
    \label{fig:cityscapesadditionalexamples}
    \end{figure}

    \Cref{fig:kittiadditionalexamples,fig:cityscapesadditionalexamples} show further qualitative
    examples of our instance segmentation on urban scenes from KITTI~\cite{KITTI2012}
    and Cityscapes~\cite{Cordts2015}. It can be seen that our approach can segment even
    high numbers of instances despite heavy occlusions and clutter.

    \section{Depth Ranges}
    As mentioned in \cref{subsec:fcnoutputrepresentation}, we discretized continuous instance depths
    into \num{19} depth classes. Instead of equidistantly splitting them, we
    chose the ranges for each class such that the sizes of objects within each depth
    class are similar. We found this option to yield slightly better results, since the
    subsequent template matching is based
    on our FCN's depth prediction and equal object sizes per depth class result in
    more reliable template matching scores.

    We defined the values as in \cref{tab:depthRanges} to provide a good
    trade-off between number of depth classes and depth resolution, as well
    as number of samples per depth class in the training data.
    As the Cityscapes dataset contains a lot of object instances labeled
    for very high distances of over \num{200} meters~\cite{Cordts2015}, the depth
    ranges had to be chosen differently than for KITTI~\cite{KITTI2012}.

    \section{Class-level Evaluation}
    \label{sec:detailedevaluation}
    \subsection{Instance-level Evaluation}
    We list class-level performances of our approach
    for instance-level semantic labeling (\textit{Ours}) and the
    baseline \textit{MCG+R-CNN} \cite{Cordts2015} in \cref{tab:instanceEval}.
    Our approach has difficulties especially for semantic classes that
    are least reliably classified by our FCN, such as bus, truck, and train
    \cf \cref{tab:pixelEval,tab:pixelInstanceEval,tab:pixelConfusion}. Best results are
    achieved for cars and humans, while we outperform the proposal-based
    baseline for all other classes by large margins in all used
    metrics.

    \subsection{Pixel-level Evaluation}
    A detailed evaluation of our performance for pixel-level semantic labeling
    can be found in \cref{tab:pixelEval,tab:pixelInstanceEval,tab:pixelConfusion}.
    Even though our main focus lies on instance-level semantic
    labeling, we achieve competitive results for all classes compared to the baselines listed in \cite{Cordts2015}.
    Using the instance-aware metric $\iiou$, we even outperform most existing works
    by a few percent points for the object classes \textit{person}, \textit{car}, and \textit{bicycle}.

    The reason for a comparably low performance on the classes \textit{bus}, \textit{truck}, and \textit{train}
    becomes evident by inspecting \cref{tab:pixelEval,tab:pixelConfusion}.
    We achieve comparably low semantic labeling results on a pixel-level for these classes
    and therefore our template matching and instance generation steps perform significantly
    worse than on all other object classes.

    \setlength{\tabcolsep}{2pt}
    \ctable[
        caption = {Class-based evaluation of existing works and our approach
                   for instance-level segmentation on Cityscapes \textit{test}
                   using metrics proposed in \cite{Cordts2015}. All numbers are in percent
                   and larger is better.},
        label   = {tab:instanceEval},
        pos     = {pb!},
        width   = \textwidth
    ]{lXccccccccc}{}{
\FL
& Metric
& \rotatedlabel{person}
& \rotatedlabel{rider}
& \rotatedlabel{car}
& \rotatedlabel{truck}
& \rotatedlabel{bus}
& \rotatedlabel{train}
& \rotatedlabel{motorcycle}
& \rotatedlabel{bicycle}
& \rotatedlabel{\textbf{mean score}} \ML
MCG+R-CNN \cite{Cordts2015} & $\mapr$ &  $ 1.3$  &  $ 0.6$  &  $10.5$  &\bst{ 6.1}&\bst{ 9.7}&\bst{ 5.9}&  $ 1.7$  &  $ 0.5$  &  $ 4.6$  \NN
Ours &$\mapr$&\bst{12.5}&\bst{11.7}&\bst{22.5}&  $ 3.3$  &  $ 5.9$  &  $ 3.2$  &\bst{ 6.9}&\bst{ 5.1}&\bst{ 8.9}\ML
MCG+R-CNN \cite{Cordts2015} &$\mapr^{50\%}$&  $ 5.6$  &  $ 3.9$  &  $26.0$  &\bst{13.8}&\bst{26.3}&\bst{15.8}&  $ 8.6$  &  $ 3.1$  &  $12.9$  \NN
Ours &$\mapr^{50\%}$&\bst{31.8}&\bst{33.8}&\bst{37.8}&  $ 7.6$  &  $12.0$  &  $ 8.5$  &\bst{20.5}&\bst{17.2}&\bst{21.1}\ML
MCG+R-CNN \cite{Cordts2015} &$\mapr^{100\text{m}}$&  $ 2.6$  &  $ 1.1$  &  $17.5$  &\bst{10.6}&\bst{17.4}&\bst{ 9.2}&  $ 2.6$  &  $ 0.9$  &  $ 7.7$  \NN
Ours &$\mapr^{100\text{m}}$&\bst{24.4}&\bst{20.3}&\bst{36.4}&  $ 5.5$  &  $10.6$  &  $ 5.2$  &\bst{10.5}&\bst{ 9.2}&\bst{15.3}\ML
MCG+R-CNN \cite{Cordts2015} &$\mapr^{50\text{m}}$&  $ 2.7$  &  $ 1.1$  &  $21.2$  &\bst{14.0}&\bst{25.2}&\bst{14.2}&  $ 2.7$  &  $ 1.0$  &  $10.3$  \NN
Ours &$\mapr^{50\text{m}}$&\bst{25.0}&\bst{21.0}&\bst{40.7}&  $ 6.7$  &  $13.5$  &  $ 6.4$  &\bst{11.2}&\bst{ 9.3}&\bst{16.7}\LL
    }

    \setlength{\tabcolsep}{6pt}
    \ctable[
        caption = {Assignment of depth classes with corresponding depth
                  ranges for the two used datasets KITTI~\cite{KITTI2012}
                  and Cityscapes~\cite{Cordts2015}.},
        label   = {tab:depthRanges},
        pos     = {p},
    ]{lcccc}{}{
\FL
\multilinecell{Class\\\phantom{A}} & \multilinecell{Depth Ranges\\KITTI} & \multilinecell{Depth\\Stepsize [m]} & \multilinecell{Depth Ranges\\Cityscapes} & \multilinecell{Depth\\Stepsize [m]}\ML
1  & 0-2 m,         & 2     & 0-6 m,                & 6 \NN
2  & 2-3.5 m,       & 1.5   & 6-8 m,                & 2 \NN
3  & 3.5-5 m,       & 1.5   & 8-10 m,           & 2 \NN
4  & 5-6 m,         & 1     & 10-12 m,          & 2 \NN
5  & 6-7 m,         & 1     & 12-14 m,          & 2 \NN
6  & 7-8.5 m,       & 1.5   & 14-17 m,          & 3 \NN
7  & 8.5-10 m,      & 1.5   & 17-20 m,          & 3 \NN
8  & 10-12 m,       & 2     & 20-23 m,          & 3 \NN
9  & 12-14 m,       & 2     & 23-27 m,          & 4 \NN
10 & 14-17 m,       & 3     & 27-31 m,          & 4 \NN
11 & 17-20 m,       & 3     & 31-36 m,          & 5 \NN
12 & 20-24 m,       & 4     & 36-41 m,          & 5 \NN
13 & 24-29 m,       & 5     & 41-47 m,          & 6 \NN
14 & 29-35 m,       & 6     & 47-54 m,          & 7 \NN
15 & 35-43 m,       & 8     & 54-63 m,          & 9 \NN
16 & 43-52 m,       & 9     & 63-73 m,          & 10    \NN
17 & 52-63 m,       & 11        & 73-86 m,          & 13    \NN
18 & 63-76 m,       & 13        & 86-100 m,             & 14    \NN
19 & 76-$\infty$ m  &       & 100-$\infty$ m    &   \LL
    }

    \setlength{\tabcolsep}{1pt}
    \ctable[
        caption = {Evaluation of our class-level performance for
                   pixel-level semantic labeling on Cityscapes \textit{test}
                   using the $\iou$ metric proposed in \cite{Cordts2015}.
                   All numbers are in percent and larger is better.},
        label   = {tab:pixelEval},
        pos     = {p},
        doinside= \tiny,
        width   = \textwidth
    ]{Xcccccccccccccccccccc}{}{
\FL
& \rotatedlabel{road}
& \rotatedlabel{sidewalk}
& \rotatedlabel{building}
& \rotatedlabel{wall}
& \rotatedlabel{fence}
& \rotatedlabel{pole}
& \rotatedlabel{traffic light}
& \rotatedlabel{traffic sign}
& \rotatedlabel{vegetation}
& \rotatedlabel{terrain}
& \rotatedlabel{sky}
& \rotatedlabel{person}
& \rotatedlabel{rider}
& \rotatedlabel{car}
& \rotatedlabel{truck}
& \rotatedlabel{bus}
& \rotatedlabel{train}
& \rotatedlabel{motorcycle}
& \rotatedlabel{bicycle}
& \rotatedlabel{\textbf{mean $\iou$}} \ML
\cite{segnet} ext. & $95.6$  &  $70.1$  &  $82.8$  &  $29.9$  &  $31.9$  &  $38.1$  &  $43.1$  &  $44.6$  &  $87.3$  &  $62.3$  &  $91.7$  &  $67.3$  &  $50.7$  &  $87.9$  &  $21.7$  &  $29.0$  &  $34.7$  &  $40.5$  &  $56.6$  &  $56.1$  \NN 
\cite{segnet} basic & $96.4$  &  $73.2$  &  $84.0$  &  $28.5$  &  $29.0$  &  $35.7$  &  $39.8$  &  $45.2$  &  $87.0$  &  $63.8$  &  $91.8$  &  $62.8$  &  $42.8$  &  $89.3$  &  $38.1$  &  $43.1$  &  $44.2$  &  $35.8$  &  $51.9$  &  $57.0$  \NN 
\cite{dpn} & $96.3$  &  $71.7$  &  $86.7$  &  $43.7$  &  $31.7$  &  $29.2$  &  $35.8$  &  $47.4$  &  $88.4$  &  $63.1$  &\bst{93.9}&  $64.7$  &  $38.7$  &  $88.8$  &  $48.0$  &  $56.4$  &  $49.4$  &  $38.3$  &  $50.0$  &  $59.1$  \NN 
\cite{ZhengJayasumana2015} & $96.3$  &  $73.9$  &  $88.2$  &\bst{47.6}&  $41.3$  &  $35.2$  &  $49.5$  &  $59.7$  &  $90.6$  &  $66.1$  &  $93.5$  &  $70.4$  &  $34.7$  &  $90.1$  &  $39.2$  &  $57.5$  &  $55.4$  &  $43.9$  &  $54.6$  &  $62.5$  \NN 
\cite{ChenPapandreou2014} & $97.3$  &  $77.7$  &  $87.7$  &  $43.6$  &  $40.5$  &  $29.7$  &  $44.5$  &  $55.4$  &  $89.4$  &  $67.0$  &  $92.7$  &  $71.2$  &  $49.4$  &  $91.4$  &  $48.7$  &  $56.7$  &  $49.1$  &  $47.9$  &  $58.6$  &  $63.1$  \NN 
\cite{Papandreou2015} & $97.4$  &  $78.3$  &  $88.1$  &  $47.5$  &  $44.2$  &  $29.5$  &  $44.4$  &  $55.4$  &  $89.4$  &  $67.3$  &  $92.8$  &  $71.0$  &  $49.3$  &  $91.4$  &\bst{55.9}&\bst{66.6}&\bst{56.7}&  $48.1$  &  $58.1$  &  $64.8$  \NN 
\cite{adelaide} & $97.3$  &  $78.5$  &  $88.4$  &  $44.5$  &\bst{48.3}&  $34.1$  &  $55.5$  &  $61.7$  &  $90.1$  &\bst{69.5}&  $92.2$  &  $72.5$  &  $52.3$  &  $91.0$  &  $54.6$  &  $61.6$  &  $51.6$  &\bst{55.0}&  $63.1$  &  $66.4$  \NN 
\cite{Cordts2015} & $97.4$  &  $78.4$  &  $89.2$  &  $34.9$  &  $44.2$  &  $47.4$  &\bst{60.1}&  $65.0$  &  $91.4$  &  $69.3$  &\bst{93.9}&  $77.1$  &  $51.4$  &  $92.6$  &  $35.3$  &  $48.6$  &  $46.5$  &  $51.6$  &\bst{66.8}&  $65.3$  \NN 
\cite{Yu2016} &\bst{97.6}&\bst{79.2}&\bst{89.9}&  $37.3$  &  $47.6$  &\bst{53.2}&  $58.6$  &\bst{65.2}&\bst{91.8}&  $69.4$  &  $93.7$  &\bst{78.9}&\bst{55.0}&\bst{93.3}&  $45.5$  &  $53.4$  &  $47.7$  &  $52.2$  &  $66.0$  &\bst{67.1}\NN[1.5pt] 
Ours & $97.4$  &  $77.7$  &  $88.8$  &  $27.7$  &  $40.1$  &  $51.5$  &\bst{60.1}&  $64.7$  &  $91.1$  &  $67.6$  &  $93.5$  &  $77.7$  &  $54.2$  &  $92.4$  &  $33.7$  &  $42.0$  &  $42.5$  &  $52.5$  &  $66.5$  &  $64.3$  \LL 
    }

     \setlength{\tabcolsep}{4pt}
    \ctable[
        caption = {Confusion matrix of our method's performance for
                   pixel-level semantic labeling on Cityscapes \textit{validation}
                   using all $8$ object classes~\cite{Cordts2015}.
                   All numbers are in percent.},
        label   = {tab:pixelConfusion},
        pos     = {p},
        width   = 0.7\textwidth
    ]{Xccccccccc}{}{
\FL
& \rotatedlabel{person}
& \rotatedlabel{rider}
& \rotatedlabel{car}
& \rotatedlabel{truck}
& \rotatedlabel{bus}
& \rotatedlabel{train}
& \rotatedlabel{motorcycle}
& \rotatedlabel{bicycle}
& \rotatedlabel{\textbf{prior}} \ML
person     & 91 &  1 &  1 &  0 &  0 &  0 &  0 &  1 & 1.14 \NN
rider      & 24 & 61 &  2 &  0 &  0 &  0 &  2 &  7 & 0.19 \NN
car        &  0 &  0 & 97 &  0 &  0 &  0 &  0 &  0 & 5.70 \NN
truck      &  0 &  0 & 25 & 56 &  1 &  1 &  0 &  0 & 0.26 \NN
bus        &  0 &  0 & 14 &  4 & 67 &  2 &  0 &  0 & 0.34 \NN
train      &  0 &  0 &  5 &  0 & 16 & 42 &  0 &  0 & 0.10 \NN
motorcycle &  6 &  4 & 13 &  0 &  0 &  0 & 64 &  7 & 0.07 \NN
bicycle    &  3 &  3 &  2 &  0 &  0 &  0 &  1 & 85 & 0.62 \LL
    }

    \setlength{\tabcolsep}{4pt}
    \ctable[
        caption = {Class-level evaluation of our object-related performance
                   for semantic segmentation on Cityscapes \textit{test}
                   using the $\iiou$ metric proposed in \cite{Cordts2015}.
                   All numbers are in percent and larger is better.\vspace{1em}},
        label   = {tab:pixelInstanceEval},
        pos     = {p},
        width   = 0.9\textwidth
    ]{Xccccccccc}{}{
\FL
& \rotatedlabel{person}
& \rotatedlabel{rider}
& \rotatedlabel{car}
& \rotatedlabel{truck}
& \rotatedlabel{bus}
& \rotatedlabel{train}
& \rotatedlabel{motorcycle}
& \rotatedlabel{bicycle}
& \rotatedlabel{\textbf{mean $\iiou$}} \ML
\cite{segnet} ext.  &  $49.9$  &  $27.1$  &  $81.1$  &  $15.3$  &  $23.7$  &  $18.5$  &  $19.6$  &  $38.4$  &  $34.2$  \NN
\cite{segnet} basic &  $44.3$  &  $22.7$  &  $78.4$  &  $16.1$  &  $24.3$  &  $20.7$  &  $15.8$  &  $33.6$  &  $32.0$  \NN
\cite{dpn} &  $38.9$  &  $12.8$  &  $78.6$  &  $13.4$  &  $24.0$  &  $19.2$  &  $10.7$  &  $27.2$  &  $28.1$  \NN
\cite{ZhengJayasumana2015} &  $50.6$  &  $17.8$  &  $81.1$  &  $18.0$  &  $25.0$  &  $30.3$  &  $22.3$  &  $30.1$  &  $34.4$  \NN
\cite{ChenPapandreou2014} &  $40.5$  &  $23.3$  &  $78.8$  &  $20.3$  &  $31.9$  &  $24.8$  &  $21.1$  &  $35.2$  &  $34.5$  \NN
\cite{Papandreou2015} &  $40.7$  &  $23.1$  &  $78.6$  &  $21.4$  &  $32.4$  &  $27.6$  &  $20.8$  &  $34.6$  &  $34.9$  \NN
\cite{adelaide} &  $56.2$  &\bst{38.0}&  $77.1$  &\bst{34.0}&\bst{47.0}&\bst{33.4}&\bst{38.1}&  $49.9$  &\bst{46.7}\NN
\cite{Cordts2015} &  $55.9$  &  $33.4$  &  $83.9$  &  $22.2$  &  $30.8$  &  $26.7$  &  $31.1$  &  $49.6$  &  $41.7$  \NN
\cite{Yu2016} &  $56.3$  &  $34.5$  &  $85.8$  &  $21.8$  &  $32.7$  &  $27.6$  &  $28.0$  &  $49.1$  &  $42.0$  \NN[3pt]
Ours &\bst{60.6}&  $33.4$  &\bst{86.7}&  $19.5$  &  $25.6$  &  $25.8$  &  $30.5$  &\bst{50.5}&  $41.6$  \LL
    }

\end{document}